\documentclass[a4paper,fleqn]{cas-dc}

\usepackage[numbers]{natbib}
\usepackage{subfigure}
\usepackage{enumitem}

\def\tsc#1{\csdef{#1}{\textsc{\lowercase{#1}}\xspace}}
\tsc{WGM}
\tsc{QE}
\tsc{EP}
\tsc{PMS}
\tsc{BEC}
\tsc{DE}
\begin{document}
\let\WriteBookmarks\relax
\def\floatpagepagefraction{1}
\def\textpagefraction{.001}

\shorttitle{Transfer Learning for Fault Diagnosis}

\shortauthors{F. M. Shakiba et~al.}

\title [mode = title]{Transfer Learning for Fault Diagnosis of Transmission Lines}

\author{Fatemeh Mohammadi Shakiba}
% [orcid=0000-0003-3631-1014]

%Corresponding author indication
 \ead{fm298@njit.edu}
 %\cormark[1]
  \address{New Jersey Institute of Technology, Newark, NJ 07102, USA}
%   city={Newark},
%     %  % citysep={}, % Uncomment if no comma needed between city and 
%     state={NJ},
%     postcode={07102},
%     country={USA}}

\author{Milad Shojaee}[orcid=0000-0002-5465-3213]
 \ead{ms2892@njit.edu}

 \author{S. Mohsen Azizi}[orcid=0000-0002-8178-2520]
 \ead{azizi@njit.edu}
%\ead[URL]{www.sayahna.org}

%\credit{Data curation, Writing - Original draft preparation}

%\affiliation[2]{organization={Sayahna Foundation},
%    city={Jagathy},
%    % citysep={}, % Uncomment if no comma needed between city and postcode
%    postcode={695014}, 
%    state={Trivandrum},
%    country={India}}

 \author{Mengchu Zhou}[orcid=0000-0002-5408-8752]
%\cormark[2]
%\fnmark[1,3]
 \ead{mengchu.zhou@njit.edu}
%\ead[URL]{www.stmdocs.in}

% \cortext[cor1]{Corresponding author}
%\cortext[cor2]{Principal corresponding author}

%\fntext[fn1]{This is the first author footnote. but is common to third
%  author as well.}
%\fntext[fn2]{Another author footnote, this is a very long footnote and
%  it should be a really long footnote. But this footnote is not yet
%  sufficiently long enough to make two lines of footnote text.}

%\nonumnote{This note has no numbers. In this work we demonstrate $a_b$
%  the formation Y\_1 of a new type of polariton on the interface
%  between a cuprous oxide slab and a polystyrene micro-sphere placed
%  on the slab.
%  }
\maketitle

\begin{abstract}
Recent artificial intelligence-based methods have shown great promise in the use of neural networks for real-time sensing and detection of transmission line faults and estimation of their locations. The expansion of power systems including transmission lines with various lengths have made a fault detection, classification, and location estimation process more challenging. Transmission line datasets are stream data which are continuously collected by various sensors and hence, require generalized and fast fault diagnosis approaches. Newly collected datasets including voltages and currents might not have enough and accurate labels (fault and no fault) that are useful to train neural networks. 
In this paper, a novel transfer learning framework based on a pre-trained LeNet-5 convolutional neural network is proposed. This method is able to diagnose faults for different transmission line lengths and impedances by transferring the knowledge from a source convolutional neural network to predict a dissimilar target dataset. By transferring this knowledge, faults from various transmission lines, without having enough labels, can be diagnosed faster and more efficiently compared to the existing methods. To prove the feasibility and effectiveness of this methodology, seven different datasets that include various lengths of transmission lines are used. The robustness of the proposed methodology against generator voltage fluctuation, variation in fault distance, fault inception angle, fault resistance, and phase difference between the two generators are well shown, thus proving its practical values in the fault diagnosis of transmission lines.
\end{abstract}

% \includegraphics{figs/grabs.pdf}


%\item Research highlights item 1
%\item Research highlights item 2
%\item Research highlights item 3

% Keywords

\section{Introduction}

Transmission lines (TLs) are exposed to severe atmospheric and environmental conditions such as lightning strikes, icing, tree interference, bird nesting, breakage, hurricanes, and aging, as well as human activities and lack of preservation and caring.
Faults cause interruption in the power flow and deteriorate the efficiency of power systems. Therefore, minimizing the effect of faults is a major goal in transmission lines. To provide end-users  with  uninterrupted electrical power, a highly reliable, fast, and accurate fault detection/classification and location estimation scheme is needed \cite{hare2016fault,BJELIC2022107825,KHOSHBOUY2022107826}.

There are several sensed physical parameters in a TL that are not constant including generator voltages, phase difference between two generators, fault resistance, fault inception angle, fault location, and length of the TL. To consider these variations, artificial intelligence methods need to be developed to deal with TL datasets that include all these parameter variations \cite{ZHENG2022107658}. 

Different lengths of TLs result in different impedances and hence different features for performance analysis. In general, TLs are divided into three categories based on their length: short TLs with maximum 50 $(km)$ length, medium TLs with 50-150 $(km)$ length, and long TLs with at least 150 $(km)$ length. In this study, to detect, classify, and locate the faults in these three categories, a general solution is proposed for the first time that is compatible with different transmission lines and avoids time-consuming trainings and high computational expenses. This work aims to show how a model's training time can be greatly reduced via transfer learning.

Neural networks (NNs) have been extensively used for TL fault diagnosis. A powerful technique to handle predictive modeling for different but somehow related problems is transfer learning, in which partial or complete knowledge of one Convolutional Neural Network (CNN) is transferred and reused to increase the speed of the training process and improve the performance of another CNN. In this study, a transfer learning-based CNN is used for the first time to detect, identify, and locate the faults for various lengths of TLs. It will be shown in this work that this methodology is reusable, fast, and accurate.

There are several studies that use CNNs for fault diagnosis of TLs. 
In \cite{fahim2020self}, a self-attention CNN framework and a time series image-based feature extraction model are presented for fault detection and classification of TLs with length of 100 $(km)$ using a discrete wavelet transform (DWT) for denoising the faulty voltage and current signals.
In \cite{rai2020fault}, authors present a customized CNN for fault detection and classification of 50 $(km)$ TLs integrated with distributed generators.
The work done in \cite{fuada2020high} proposes a machine learning-based CNN for TLs with length of 280 $(km)$ to perform fault detection and classification using DWT for feature extraction.

Shiddieqy et al. \cite{shiddieqy2019power} present another methodology that considers all features of the TL faults to generate various models for robust fault detection. Then, they take advantage of various artificial intelligence methods including CNN to achieve a $100\%$ detection accuracy. The length of the TLs in this study is 300 $(km)$.
The study in \cite{chen2016detection} presents a scheme to detect and categorize faults in power TLs with length of 200 $(km)$ using convolutional sparse auto-encoders. This approach has the capability to learn extracted features from the dataset of voltage and current signals, automatically, for fault detection and classification. To generate feature vectors, convolutional feature mapping and mean pooling methods are applied to multi-channel signal segments.

There are also several studies that apply transfer learning methods to time series datasets in different applications.
Fawaz et al. \cite{fawaz2018transfer} show how to transfer deep CNN knowledge for time series dataset classification. In \cite{guo2018deep}, an intelligent method is proposed as a deep convolutional transfer learning network which diagnoses the dynamic system faults using an unlabeled dataset. In \cite{shao2018highly}, a deep learning framework is presented to achieve accurate fault diagnosis using
transfer learning which speeds up the training of the deep CNN. Shao et al. \cite{shao2020intelligent} propose an intelligent fault diagnosis method for a rotor-bearing system which is based on a modified CNN with transfer learning. The study performed in \cite{xu2019digital} presents
a two-phase digital-twin-assisted fault diagnosis method which takes advantage of deep transfer learning. It performs fault diagnosis in the development and maintenance phases. Li et al. \cite{li2020deep} presented a deep adversarial transfer learning network  to diagnose new and unlabeled emerging faults in rotary machines.
The contributions of this work are as follows:
\begin{enumerate}
\item Proposing a generalized approach for various TLs' fault detection and location estimation using transfer learning technique for the first time ; and
\item Comparing the proposed methodology with three benchmarks and showing its effectiveness in prediction performance.
\end{enumerate}

The rest of the paper is organized as follows. Section \ref{overview} presents preliminary concepts used in this study including time series classification, CNNs, and transfer learning technique. In Section \ref{TL model}, the TL model used in this study as well as the feature generation approach are discussed. Section \ref{proposed_method} describes the dataset generation procedure and the proposed transfer learning-based method. Section \ref{results} discusses the results and compares the proposed method with three benchmarks including K-means clustering, CNN without transfer learning, CNN with transfer learning technique but without fine tuning process. Finally, conclusions are made in Section \ref{conclusion}.

\section{Preliminaries} \label{overview}

In this section, an overview of time series data classification, CNNs, and transfer learning procedure is provided. 

\subsection{Time Series Classification}
In general, time series data can be defined in two different ways as described below \cite{fawaz2018transfer}.
\begin{itemize}
    \item Definition 1. A time series data is an ordered (time dependant) set of real values such as $X = [x_1, x_2,...,x_n ]$ where $n$ is the number of real values and the length of $X$ \cite{fawaz2018transfer}.
    \item Definition 2. A time series dataset $D$ is defined as $D=\{(x_1, y_1),...,(x_n,y_n)\}$ with length of $n$ which consists of a collection of pairs $(x_i,y_i)$ where $x_i$ is a time series data point with its corresponding label (class) as $y_i$ \cite{fawaz2018transfer}.
\end{itemize}

Time series classification is defined as classifying the dataset $D$ by considering every input $x_i$ to train a classifier which maps the given inputs to the given labels based on every class variable $y_i$ \cite{fawaz2018transfer}. It is clear that sometimes $D$ consists of a set of pairs in which the inputs and labels are vectors.
In this study, the system utilizes voltage and current waveforms recorded from one end of a two-bus TL as inputs, and considers 10 types of faults as well as the non-faulty case as the labels (classes). Fast Fourier transform (FFT) is applied to both current and voltage waveforms to generate the amplitude of the main frequency component of the signals. Therefore, the dataset used to train the CNN includes seven different features from which the first three features are associated with voltages, the second three features are associated with currents, and the last one is related to the zero sequence signal which is the average of the phase currents. The reason for having the last feature will be discussed later in Section \ref{TL model}.
\subsection{Convolutional Neural Network}
Having the capability of learning hierarchical features independently
from inputs, CNNs have been widely used for image datasets. CNNs have a minimum need for pre-processed data and can handle high dimensional datasets faster and with more details in comparison with most of artificial CNNs \cite{hubel1962receptive,kiruthika2020classification}. In general, CNNs include three types of layer:
convolution layer, pooling layer, and fully connected layer.
Convolutional and pooling layers incorporate convolution blocks which are stacked for feature extraction purpose. Fully connected layers are used as classifiers and the output layer, which is a fully connected one, performs the classification or regression task.

The main advantages of CNN architecture are local receptive fields, shared weights, and the pooling operation. CNNs take advantage of the concept of a local
receptive field which means that only a small focused area of the input data is connected to each node in a convolution layer. Because of this trait, the number of parameters is reduced considerably in CNN which in turn decreases the training computational expenses of the NN \cite{shao2018highly}.

Numerous studies have been performed in TL fault detection, classification, and localization problems using CNNs to achieve higher accuracies. These studies are divided into two different categories: The methodologies with a focus on image-based datasets recorded from outdoor TLs \cite{lei2019intelligent,wang2019image,dai2020fast}, and the ones that consider time-series voltage and current waveforms recorded from generators and are fed to CNNs as blocks of data points. The methodology proposed in this study belongs to the second category.

\subsection{Transfer Learning}
Transfer learning is proposed to solve the learning problems between two or multiple domains. The combination of deep learning and
transfer learning shows notable improvements in the accuracy and time of fault diagnosis approaches. Transfer learning consists of two steps: First, the process of training a source neural network on a source dataset and task, and second, transferring the learned features and knowledge to a new network to help the training process of the new related (target) dataset.
Two main concepts are used in transfer learning, namely, domain and task which are defined below.
\begin{itemize}
\item Definition 3 (Domain \cite{pan2009survey,zheng2019cross}). A domain $\mathcal{D}=\{ \mathcal{X},\mathcal{P}(X) \}$ consists of two elements, namely, a feature space $\mathcal{X}$ and a marginal probability distribution $\mathcal{P}(X)$ where $X=\{x_i\}_{i=1}^n \in \mathcal{X}$ is a dataset in which every ${x_i} \in \mathbb{R}^D$ is sampled from this domain.

\item Definition 4 (Task \cite{pan2009survey,zheng2019cross}). A task $\mathcal{T}=\{ \mathcal{Y}, \mathcal{P}(Y|X) \}$ consists of two elements given a domain\\ $\mathcal{D}=\{\mathcal{X},\mathcal{P}(X)\}$, where $\mathcal{Y}$ stands for the label space and $\mathcal{P}(Y|X)$ is the conditional probability distribution in which $Y=\{y_i\}_{i=1}^n$ shows the label vector of $X$ with $y_i\in \mathcal{Y}$ as the label of $x_i$.
\end{itemize}
There are two domains in transfer learning, namely, a source domain $\mathcal{D}_s=\{\mathcal{X}_s,\mathcal{P}_s({X_s})\}$ and a target domain $\mathcal{D}_t=\{\mathcal{X}_t,\mathcal{P}_t(X_t)\}$ where $\mathcal{X}_s$ and $\mathcal{X}_t$ show the feature spaces of the source and target domains, respectively, and $\mathcal{P}_s(X_s)$ and $\mathcal{P}_t(X_t)$ stand for the marginal probability distribution of them.
Based on definitions 3 and 4, the definition of transfer learning is presented below.
\begin{itemize}
    \item Definition 5 (Transfer Learning \cite{zheng2019cross}). Considering the source domain $\mathcal{D}_s$, learning task $\mathcal{T}_s$, target domain $\mathcal{T}_s$, and learning task $\mathcal{T}_t$, the goal of transfer learning is to promote the performance of target predictive function $f_t(.)$ in $\mathcal{D}_t$ by inducing the knowledge to $\mathcal{D}_s$ and $\mathcal{T}_s$ while $\mathcal{D}_s\neq\mathcal{D}_t$ or $\mathcal{T}_s\neq\mathcal{T}_t$.
\end{itemize}
In the case of $\mathcal{T}_s = \mathcal{T}_t$, a common subproblem of transfer learning takes place which is called domain adoption.
In this study, the relationship between the source and target data is the domain adoption because the seven aforementioned features (currents, voltages, and zero sequence current) repeat in both the source and target TLs which only differ in length \cite{csurka2017domain}.

Transfer learning provides the ability to distribute learned features across different learning applications. The focus of transfer learning is on the deep learning training step to enhance its capability in common feature extraction and adoption among multiple datasets of a similar problem. Transfer learning is based on using the pretrained layers on a source task to solve a target task. For this purpose, a pretrained model in which its fully connected layers are cut off is used and its convolutional and pooling layers become frozen (their weights are not updated) to perform the role of feature extractors. To adjust the given pretrained network with the target dataset and efficient target classification, the fully connected layers (classifier section) need to update their weights.
\begin{figure}[htbp]
\vspace{-2mm}
\begin{center}
\includegraphics[scale=0.345]{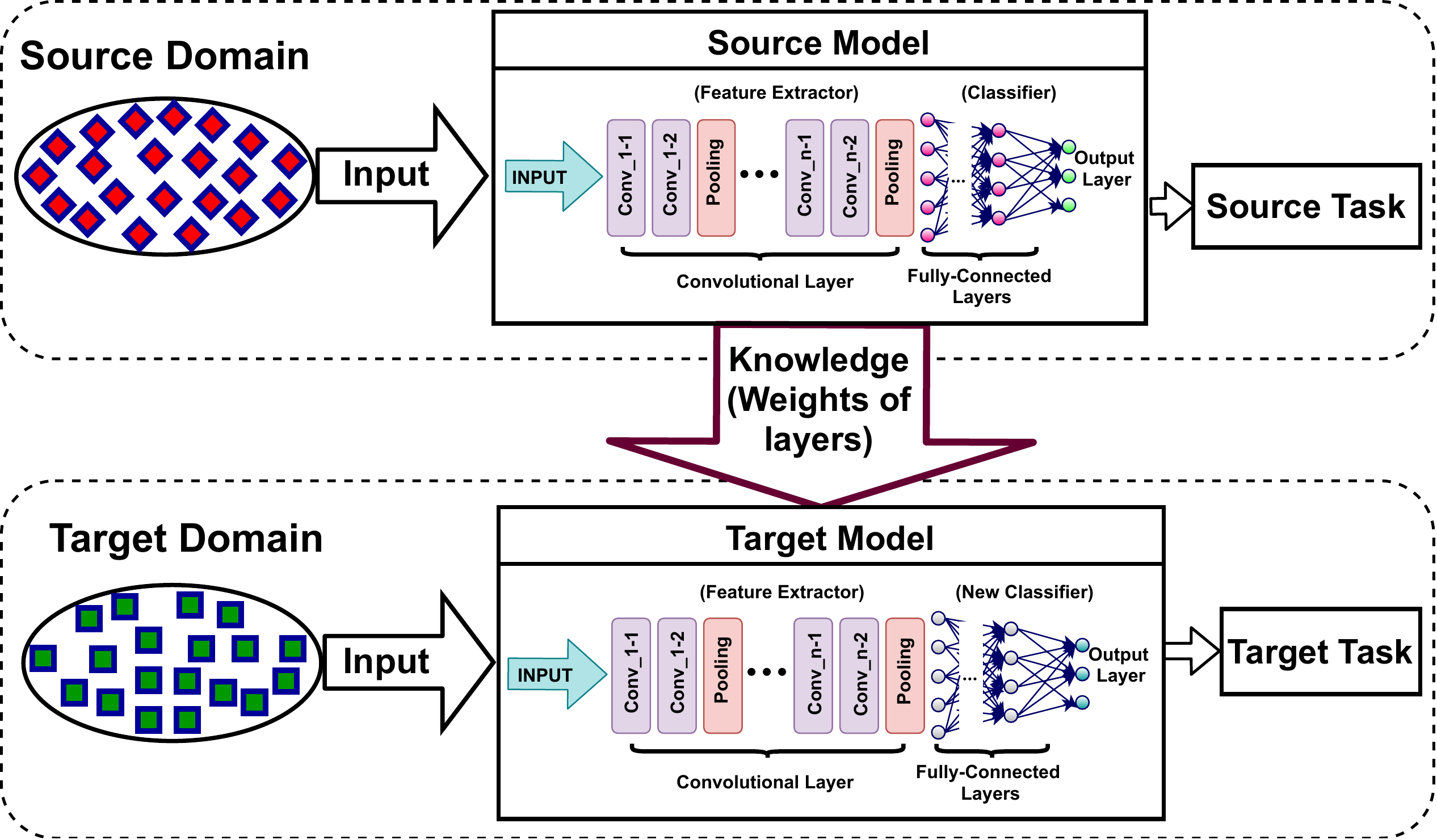}
\caption{A typical model of transfer learning.}
\label{fig:tl}
\end{center}
\vspace{-3mm}
\end{figure}

There are two different approaches in transfer learning including feature extraction and fine tuning. In the first approach, the fully connected layers are completely removed and based on the target dataset, the new fully connected layers are integrated with the frozen pretrained layers. In the fine tuning process, the structure of fully connected layers from the pretrained model is saved and only their weights are updated. Fig. \ref{fig:tl} shows the basic concept of transfer learning representing two different datasets which have similarities. These datasets are fed into source and target models. The knowledge is transferred to the target model to perform the training of target task faster.
In this study, the fine tuning approach is used to achieve satisfactory results. Transfer learning method is used in this study for TL fault diagnosis problems to detect the rare occurrence of failures which are difficult or impossible to be labeled. It should be noted that a transfer learning-based CNN has not been used before for TL fault diagnosis problems. 

\section{Transmission Line Model and Feature Generation}\label{TL model}
\begin{figure}[htbp]
\begin{center}
\includegraphics[scale=0.35]{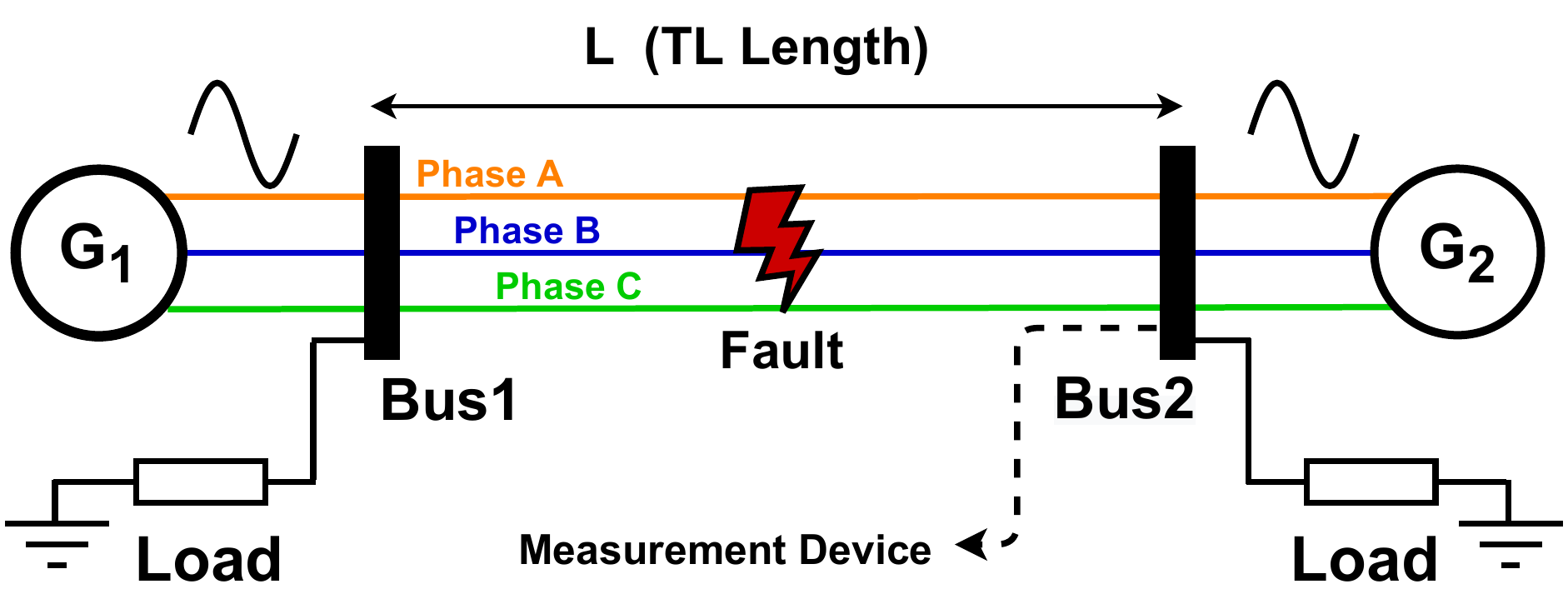}
\caption{A three-phase and two-generator power system.}
\label{fig:tlmodel}
\end{center}
\vspace{-4mm}
\end{figure}

In this study, a power system with two generators which are connected through a 100 $(km)$ three-phase TL is used. Based on some recent studies\cite{udofia2020fault,abdel2016detection,mahmud2018robust,fahim2020self}, a common length is chosen which stands in the medium range of transmission line lengths and its multiplications can reside in short or long TLs. The voltage of both generators is 240 $(kV)$ and their frequency is 60 ($Hz$) as shown in Fig. \ref{fig:tlmodel}. This model is simulated in MATLAB Simulink’s Simscape Power System, and all the ten short-circuit faults i.e., single Line-to-Ground (LG), Line-to-Line (LL), double Line-to-Ground (LLG), and all-Lines-connected (-to-Ground) (LLL/LLLG) as well as no-fault state are taken into account. 
%%\vspace{-4mm}

The parameters of this TL model are shown in Table \ref{tab:Param}, and Table \ref{tab:Source} presents the features of generators. The model studied in this paper is based on \emph{IEEE 39}-Bus System which includes 10 generators and 46 lines.
\vspace{3mm}

\begin{table}[h]
\centering
\caption{TL nominal parameters.}
\begin{tabular}{|c|cc|}
\hline 
Parameter & Zero Sequence & Positive Sequence\\ \hline 
R ($\Omega$/$(km)$)  & $0.3864$ & $0.01273$\\ 
L ($mH$/$(km)$) & $4.1264$ & $0.9337$ \\ 
C ($\mu$F/$(km)$)  & $7.751\times 10^{-3}$ & $12.74\times 10^{-3}$ \\ \hline 
\end{tabular}
\label{tab:Param}
\end{table}
\vspace{-3mm}
\begin{table}[h]
\centering
\caption{Source and load nominal parameters.}
\begin{tabular}{|c|ccc|}
\hline
Nominal Parameter & Source 1 & Source 2 & Load \\ \hline 
{Phase to Phase Voltage ($kV$)} & $240$ & $240$ & $240$\\ 
Frequency ($Hz$) & $60$ & $60$ & $60$ \\ 
Resistance ($\Omega$) & $0.08929$ & $0.08929$ & ---\\ 
Inductance ($mH$) & $16.58$ & $16.58$ & ---\\ 
Active Power ($kW$) & --- & --- & $100$\\ 
{Inductive Reactive Power (\emph{kVAR})} & --- & --- & $<100$\\ 
{Capacitive Reactive Power (\emph{kVAR})} & --- & --- & $<100$\\  \hline
\end{tabular}
\label{tab:Source}
\end{table}

The general datasets used in this study consist of the amplitudes of the main harmonics of the voltage and current waveforms calculated by FFT, which computes the frequency components faster and more efficiently than the conventional Fourier transform. For this purpose, after each fault incident, 1.5 cycles of the voltage and current waveforms are selected. Then, their main harmonics are computed, normalized and fed into the fault diagnosis module. 
\begin{figure*}[t]
\begin{center}
\includegraphics[scale=0.40]{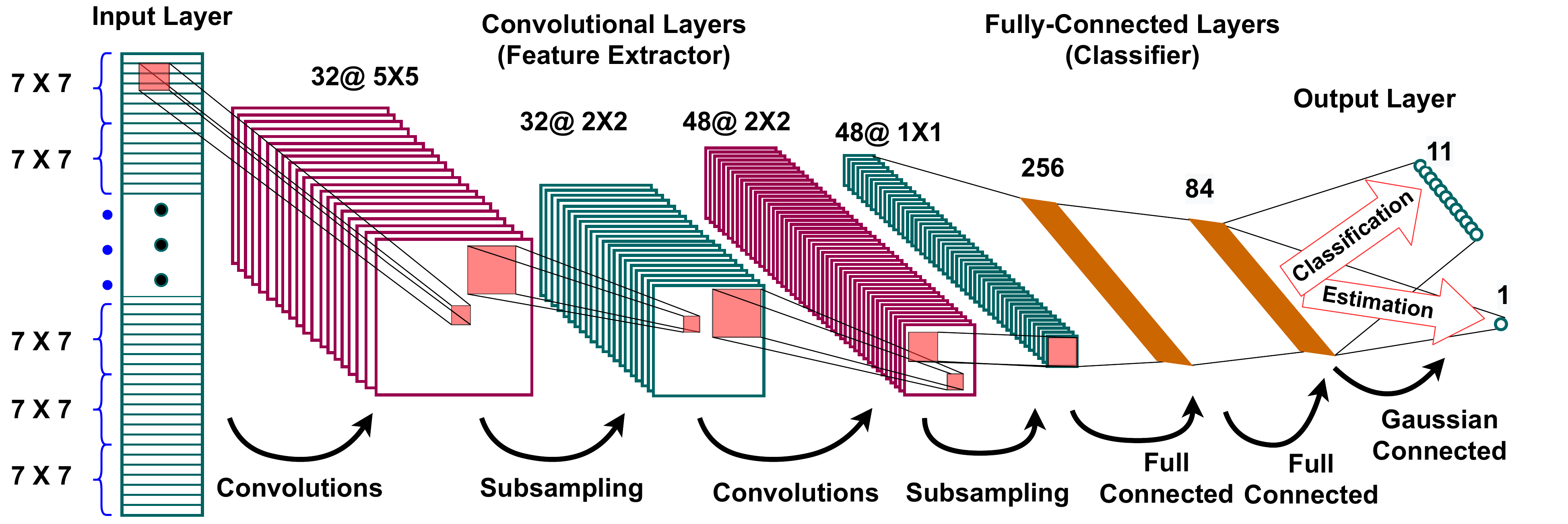}
\caption{The structure of LeNet-5.}
\label{fig:lenet5}
\end{center}
\vspace{-5mm}
\end{figure*}
To assess the robustness of the proposed methodology, some variations in the TL model are considered such as fault type, location, inception angle, resistance, voltage amplitudes of the generators, and the phase difference between them. Such variations help the system to generate datasets that are large enough to generate reliable results. In general, the dataset used in this work includes 7 features that are 3 phase voltages, 3 phase currents, and a zero-sequence current which is a detector for ground faults. In other words, a fault diagnosis model cannot distinguish between LL and LLG faults only by considering phase voltage and current signal values. In such cases, a zero-sequence current, which is the average of the phase currents, is considered to detect ground faults.
\begin{table}[H] 
\vspace{-2mm}
\centering
\caption{Parameter values for the generation of training dataset.}
\begin{tabular}{|c|c|}
\hline 
\textbf{Parameter} &\textbf{Variations}\\ \hline \rule{0pt}{1\normalbaselineskip}
{ Fault Distance $(km)$}  &\begin{tabular}[c]{@{}c@{}} {1.2,10, 24, 40, 60, 95} \end{tabular}\\ 
{Fault Inception angle ($^\circ$)} & \begin{tabular}[c]{@{}c@{}}{$1,20$,$50,100,150$}\end{tabular} \\ 
{Fault Resistance $R_f$ ($\Omega$)}  &\hspace{1.5 pt} \begin{tabular}[c]{@{}c@{}}{$0.1,1,10,20,30,40,50,60$} \end{tabular}\hspace{1.5 pt} \\
{Phase Difference$\Delta \phi$ ($^{\circ}$)} & $-30,0,30$ \\ 
\begin{tabular}[c]{@{}c@{}}{Voltage Fluctuations}\\{$\Delta$ $V_i$ ($kV$) $=V_1-V_2$} \end{tabular}& $-40,0,40$\\ \hline
\end{tabular}
\label{tab:variation}
\vspace{-2mm}
\end{table}

Table \ref{tab:variation} shows the parameters with their variations to generate robust results. It should be noted that because the length of a TL is the critical parameter in this study, the variation of fault location is defined as a dependent parameter to the TL length ($L=100 km$), and for different lengths, the initial fault distances for the reference dataset would be different as well.

\begin{table*}[ht]
\centering
\caption{The average accuracy results of K-means clustering for 11 types of faults based on statistical testing.}
\label{tab:kmeans}
\begin{tabular}{|c|c|c|c|c|c|c|c|c|}
\hline
{Length $(km)$}& 12.5 & 25 & 50 & 100 & 200 & 400 & 800 & Average\\ \hline
\begin{tabular}[c]{@{}c@{}}Accuracy (\%)\end{tabular}& 82.42 $\pm$ 0.11 & 82.47 $\pm$ 0.07 & 83.47 $\pm$ 0.15 & 85.09 $\pm$ 0.16 & 87.12 $\pm$ 0.10 & 87.67 $\pm$ 0.09 & 83.87 $\pm$ 0.17 & 84.58 $\pm$ 0.12 \\ \hline
\end{tabular}
%\vspace{-5mm}
\end{table*}

\section{Proposed Fault Diagnosis Approach}\label{proposed_method}
In this section, the proposed method based on transferred CNN is described. 
Acquiring data from one end of a TL is the initial step in fault  detection/classification and location estimation. As shown in Fig.  \ref{fig:tlmodel}, the sensing/measurement device is placed at the second bus to record the current and voltage of the three-phase TLs. This process is performed via a 30-sample time window with the sampling frequency of 1.2 ($kHz$), and FFT is applied to each window at every step to extract the features.

After normalizing the extracted features, the data points are fed to the fault diagnosis and location estimation networks. In the fault diagnosis process, 4 outputs are generated for each data point which are associated with phases A, B, C, and the ground G. When there is no fault, the value of all these 4 outputs are zero and by switching each output to one, the faulty state of that phase is detected. If not all outputs are zero, then at least two of them are one that shows the connection of those two outputs together (or to the ground). Therefore, in the output layer of the CNN, eleven states can occur including 0000 (or 1111), 0011, 0110, 1100, 1001, 0101, 1010, 0111, 1011, 1101, 1110. A parallel process using another CNN is performed for the location estimation of the faults. In this system, the input dataset is the same as that of the classification procedure; however, the outputs are the locations of the faults which are real numbers, not binary.

The detection, classification, and location estimation of TL faults are done by considering the 7-dimension input dataset divided into $7\times 7$ small blocks. The features are extracted by striding over these blocks and performing the convolution computation for each window.
For this purpose, one of the earliest pretrained CNNs called LeNet-5 \cite{lecun2015lenet} is designed by using Keras library \cite{chollet2015keras} in Python. LeNet-5 is chosen for this study because of its simple and straightforward architecture including 2 sets of convolutional and average pooling layers following by a flatten layer, 2 fully connected layers, and ultimately a Softmax classifier. To obtain high accuracy, ReLu activation function is used for all the layers except the output which takes advantage of Softmax function for classification and the linear regression function for location estimation. This architecture is depicted in Fig. \ref{fig:lenet5}. As shown in this figure, the input data points are given by $7\times7$ matrices to the network in order to emulate the image behaviors, and the output layer consists of 11 neurons (equal to the number of classes) for classification or one neuron for the location estimation tasks using CNN.

To apply the transfer learning method to the TL fault diagnosis problem, the LeNet-5 network is trained with the dataset of a TL with length $L$ which is initially equal to 100 $(km)$. Then, the weights of convolutional and pooling layers of this network are saved as ".npy" files and a new LeNet-5 with the same architecture uses these files to apply the generated weights to the corresponding layers. Therefore, the weights of convolutional and pooling layers in the new LeNet-5 are frozen and equal to the first four trained layers of the initial LeNet-5. These 4 layers perform the feature extraction, and the rest of the layers, which are free to be updated based on the new datasets (dor TLs with difeerent length of $\frac{L}{8}=12.5$, $\frac{L}{4}=25$, $\frac{L}{2}=50$, $2\times L=200$, $4\times L=400$, and $8\times L=800$), perform the classification and adaptation tasks. As it is shown in the next section, this process reduces the training time considerably as compared to the case that a new CNN is trained individually for each specific TL length.

\section{Results and Discussions}\label{results}

This section presents the results of the proposed transfer learning-based CNN method. Simulations are run on a PC with Microsoft Windows 10. This PC uses an Intel Corei7-4710MQ @ 2.50 GHz processor with 8 GB of RAM. Keras 2.3 library \cite{chollet2015keras} with TensorFlow 2.0 backend \cite{tensorflow2015-whitepaper} is used to design LeNet-5, and Scikit-learn library \cite{pedregosa2011scikit} is used for classification modules.
An LeNet-5 is used with the architecture depicted in Fig. \ref{fig:lenet5} to classify the faults and estimate their locations. 

To achieve reliable results, a statistical testing method is implemented and each experiment is performed 30 times. The values in the tables and figures in this section are all based on the thirty iterations of running each analysis \cite{haroush2021statistical}. In case that the values are different, the variations are shown in the corresponding tables.
This section is divided into two main subsections including classification and location estimation of TL faults.
\vspace{-2mm}
\subsection{Fault Classification}
In this subsection, four different simulations are performed to prove the reliability and accuracy of the proposed approach. Different classification and clustering approaches are performed including (a) K-means algorithm for clustering without knowing the labels, (b) a dedicated CNN (a NN that is independently and specifically trained for one TL) for each specific TL length, (c) transfer learning method without fine tuning, and (d) transfer learning method with fine tuning which is the main focus in this paper.

First, a K-means clustering algorithm is implemented to categorize the faults without knowing their true labels. Then, the results are compared with the true labels to calculate the accuracy which is reported in Table \ref{tab:kmeans}. The number of iterations needed for the convergence of TL length variants are between 15 to 23 in this algorithm \cite{zhao2021landslide}.

In the second step, a dedicated CNN method is used for each TL length variant. The chosen CNN is LeNet-5 which produces acceptable results and is used for the comparison with the proposed transfer learning-based method. For this purpose, 70\% and 30\% of the data are used for training and testing, respectively. Table \ref{tab:without_transfer} demonstrates the accuracy, precision, recall, F1 Score, and training time for 7 different lengths of TLs without using the transfer learning technique. The parameters of performance evaluations are defined as (\ref{eq:acc})-(\ref{eq:f1}).
%F1 Score is the weighted average of precision and recall and it is defined as (\ref{eq:f1}):
\begin{equation}
\label{eq:acc}
    Accuracy = \frac{TP+TN}{TP+TN+FP+FN}
\end{equation}
\begin{equation}
\label{eq:or}
    Precision = \frac{TP}{TP+FP}
\end{equation}
\begin{equation}
\label{eq:re}
    Recall = \frac{TP}{TP+FN}
\end{equation}
\begin{equation}
\label{eq:f1}
    F1 = 2\times\frac{Recall \times Precision}{Recall + Precision}
\end{equation}

The provided results are obtained after 64 epochs and with $3\times10^{-4}$ learning rate for the ``adam'' optimizer. 
\begin{table}[h]
\centering
\caption{Classification results of various lengths for TLs using a dedicated LeNet-5 NN.}
\label{tab:without_transfer}
\begin{tabular}{|c|c|c|c|c|c|c|c|}
\hline
\multicolumn{1}{|c|}{\begin{tabular}[c]{@{}c@{}}Length \\$(km)$\end{tabular}}& \multicolumn{1}{c|}{\begin{tabular}[c]{@{}c@{}}Accuracy\\ (\%)\end{tabular}} & \multicolumn{1}{c|}{\begin{tabular}[c]{@{}c@{}}Precision\\ (\%)\end{tabular}} & \multicolumn{1}{c|}{\begin{tabular}[c]{@{}c@{}}Recall\\ (\%)\end{tabular}} & \multicolumn{1}{c|}{\begin{tabular}[c]{@{}c@{}}F1 Score\\ (\%)\end{tabular}} & \multicolumn{1}{c|}{\begin{tabular}[c]{@{}c@{}}Training Time\\ ($sec$)\end{tabular}} \\ \hline

12.5& 99.7& 97.61& 97.54& 97.57& 3169.12\\ \hline
25& 99.69& 97.57& 97.45& 97.50& 3493.55\\ \hline
50& 99.46& 95.84& 95.58& 95.70& 3441.43\\ \hline
100 & 99.42& 95.5& 95.23& 95.36& 3167.27 \\ \hline
200& 99.15& 93.38& 93.08& 93.22& 2783.07\\ \hline
400& 99.1& 93.08& 92.56& 92.81& 3217.381\\ \hline
800& 98.62& 89.85& 87.83& 88.82& 3168.63\\ \hline
\end{tabular}
%\vspace{-2mm}
\end{table}

In the third step, a transfer learning-based method without fine tuning process with 64 epochs is used. In other words, all layers of LeNet-5 hold on to the weights of pretrained layers and become frozen. The purpose of this step is to show the effect of fine tuning process on the result of TL fault classification. Table \ref{tab:acc_nofine} shows the results of this step.

\begin{table}[h]
%\vspace{-3mm}
\center
\caption{Classification results for various lengths of TLs using LeNet-5-based transfer learning method \underline{without} fine tuning process.}
\label{tab:acc_nofine}
\begin{tabular}{|c|c|c|c|c|c|}
\hline
\begin{tabular}[c]{@{}c@{}}Length\\ $(km)$\end{tabular} & \begin{tabular}[c]{@{}c@{}}Accuracy\\ (\%)\end{tabular} & \begin{tabular}[c]{@{}c@{}}Precision\\ (\%)\end{tabular} & \begin{tabular}[c]{@{}c@{}}Recall\\ (\%)\end{tabular} & \begin{tabular}[c]{@{}c@{}}F1 Score\\ (\%)\end{tabular} & \begin{tabular}[c]{@{}c@{}}Training Time \\($sec$)\end{tabular} \\ \hline
12.5& 90.80& 74.06& 50.59& 60.11& 1101.04\\ \hline 
25& 90.96& 75.3& 52.62& 61.94& 1111.66\\ \hline
50& 90.89& 72.49& 55.37& 62.78& 1103.21\\ \hline 
200& 91.19& 74.25& 59.7& 66.18& 1123.83\\ \hline 
400& 90.90& 71.99& 56.18& 63.10& 1195.01\\ \hline 
800& 90.11& 65.07& 42.35& 51.30& 1193.73\\ \hline
\end{tabular}
%\vspace{-2mm}

\end{table}

In the fourth step, the proposed method is implemented using transfer learning method and LeNet-5 with fine tuning approach. Table \ref{tab:with_transfer} demonstrates the fault classification results of the transfer learning method for various lengths of TLs. In order to perform a fair comparison among the transfer learning-based and non-transfer learning-based methods, the number of epochs ($64$) and learning rate ($3\times10^{-4}$) for ``adam'' optimizer) remain the same in all of them. 

\begin{figure}[h]
\begin{center}
\includegraphics[scale=0.7]{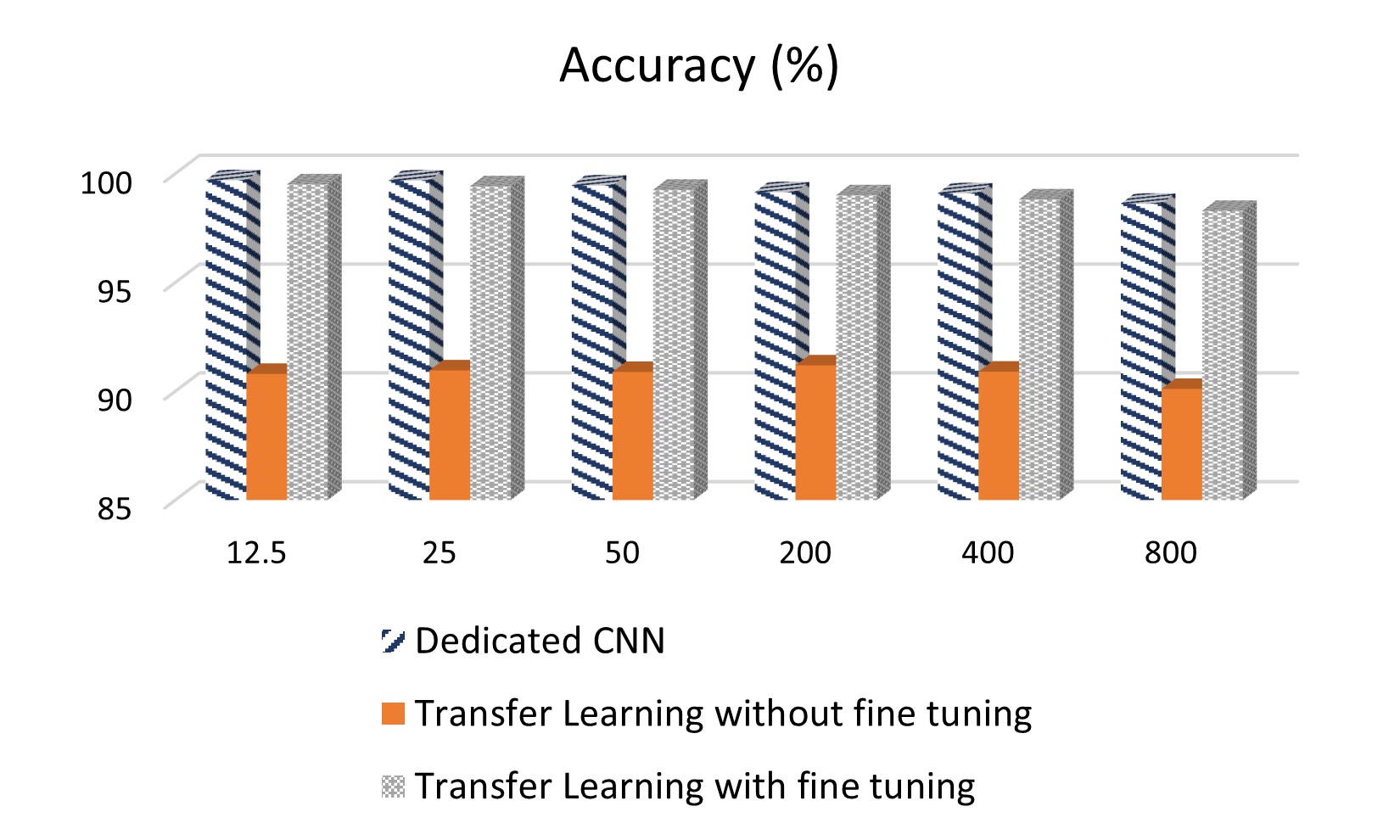}
\caption{Accuracy comparison between three methods including using the transfer learning method with and without fine tuning process, and dedicated CNN for classification of TL faults.}
\label{fig:acc_comp}
\end{center}
\vspace{-4mm}
\end{figure}

\begin{figure}[h]
%\vspace{-5mm}
\begin{center}
\includegraphics[scale=0.7]{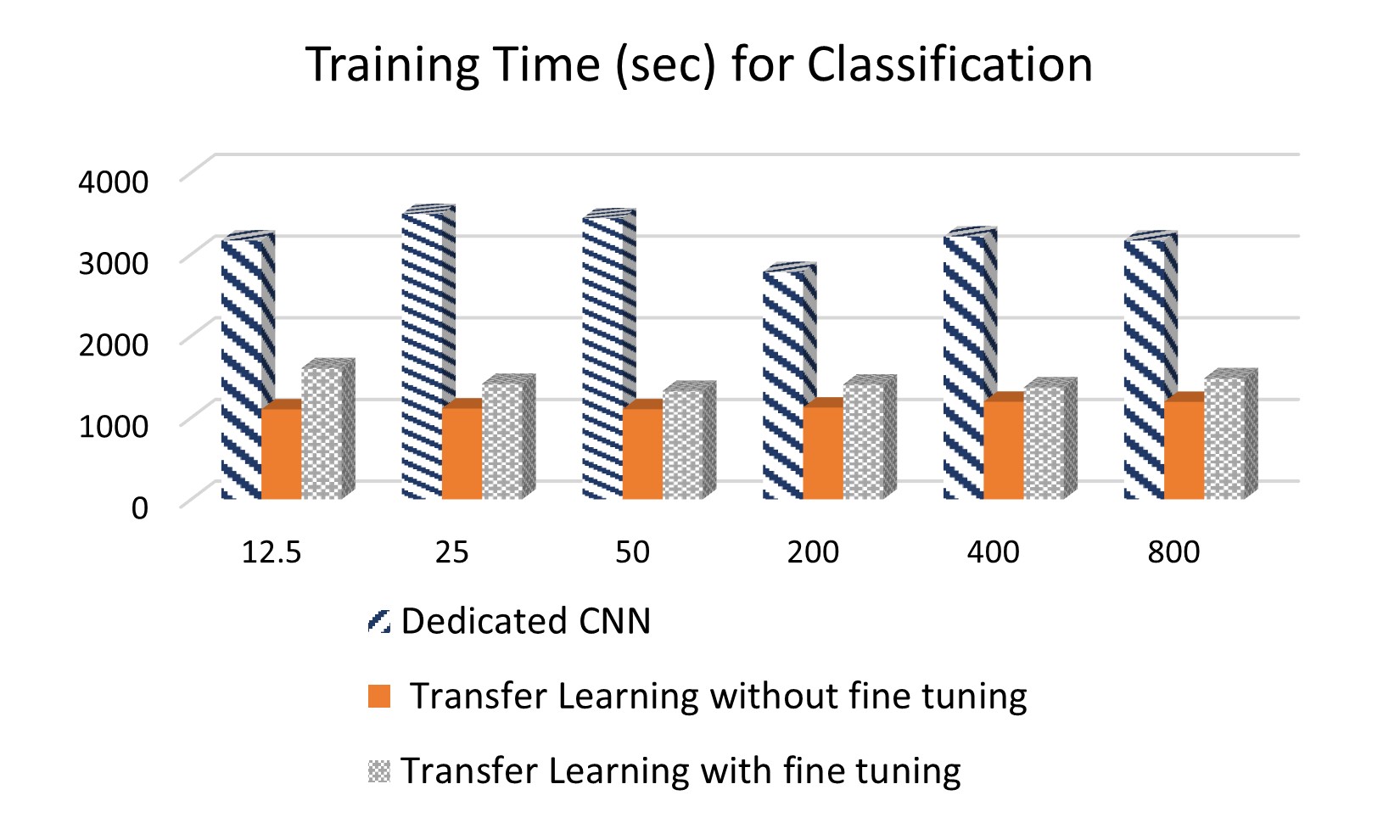}
\caption{Training time comparison between three methods of using the transfer learning  with and without fine tuning process, and the dedicated CNN for classification of TL faults.}
\label{fig:train_classification}
\end{center}
\vspace{-2mm}
\end{figure}

According to the results given in Tables \ref{tab:without_transfer} and \ref{tab:with_transfer}, the training time of the transfer learning-based method is almost \underline{half} of the training time of the dedicated CNN methodology (which is a LeNet-5 NN without transfer learning and fine tuning), while the accuracy values are almost similar. This result is achieved with negligible loss in accuracy level. As it is shown in Fig. \ref{fig:acc_comp}, the difference between the proposed transfer learning-based method accuracy and the dedicated CNN method is less than 0.5 \% for all various lengths of TLs. Fig. \ref{fig:train_classification} shows the comparison of the training time between the transfer learning-based and non-transfer learning-based approaches. As it is clear, transfer learning decreases the training time of the classification to less than \underline{half} of the training time of the dedicated CNN approach. The k-means clustering technique generates much lower accuracy results (in average 85\%) which is another proof for the significance and reliability of using transfer learning method when labels are missing or inadequate.
\begin{table}[h]
\center
\caption{Classification results for various lengths of TLs using LeNet-5-based transfer learning method \underline{with} fine tuning process.}
\label{tab:with_transfer}
\begin{tabular}{|c|c|c|c|c|c|}
\hline
\multicolumn{1}{|c|}{\begin{tabular}[c]{@{}c@{}}Length \\$(km)$\end{tabular}} & \begin{tabular}[c]{@{}c@{}}Accuracy\\ (\%)\end{tabular} & \begin{tabular}[c]{@{}c@{}}Precision\\ (\%)\end{tabular} & \begin{tabular}[c]{@{}c@{}}Recall\\ (\%)\end{tabular} & \begin{tabular}[c]{@{}c@{}}F1 Score\\ (\%)\end{tabular} & \begin{tabular}[c]{@{}c@{}}Training \\Time($sec$)\end{tabular} \\\hline %\begin{tabular}[c]{@{}c@{}}Test time\\ per Fault\\ ($\mu s$)\end{tabular} \\ 
12.5& 99.48$\pm$0.04& 95.9$\pm$0.3& 96.0$\pm$0.3& 96.0$\pm$0.3& 1307$\pm$46\\ \hline
25& 99.36$\pm$0.04& 95.5$\pm$0.3& 95.1$\pm$0.3& 95.3$\pm$0.3& 1414$\pm$46\\ \hline
50&99.26$\pm$0.03 & 94.3$\pm$0.2& 93.8$\pm$0.3& 94.0$\pm$0.2& 1328$\pm$49\\ \hline  
200& 99.01$\pm$0.03& 92.2$\pm$0.2  & 91.6$\pm$0.2& 91.9$\pm$0.2& 1406$\pm$43\\ \hline  
400& 98.82$\pm$0.04& 91.2$\pm$0.2& 89.8$\pm$0.3& 90.5$\pm$0.2& 1371$\pm$39\\ \hline 
800& 98.29$\pm$0.05 & 87.4$\pm$0.3& 80.9$\pm$0.4& 84.1$\pm$0.3& 1483$\pm$47\\  \hline
\end{tabular}
%\vspace{-4mm}
\end{table}

%\vspace{-4mm}

%\vspace{-3mm}
\begin{figure}[htbp]
\vspace{-3mm}
\begin{center}
\includegraphics[scale=0.7]{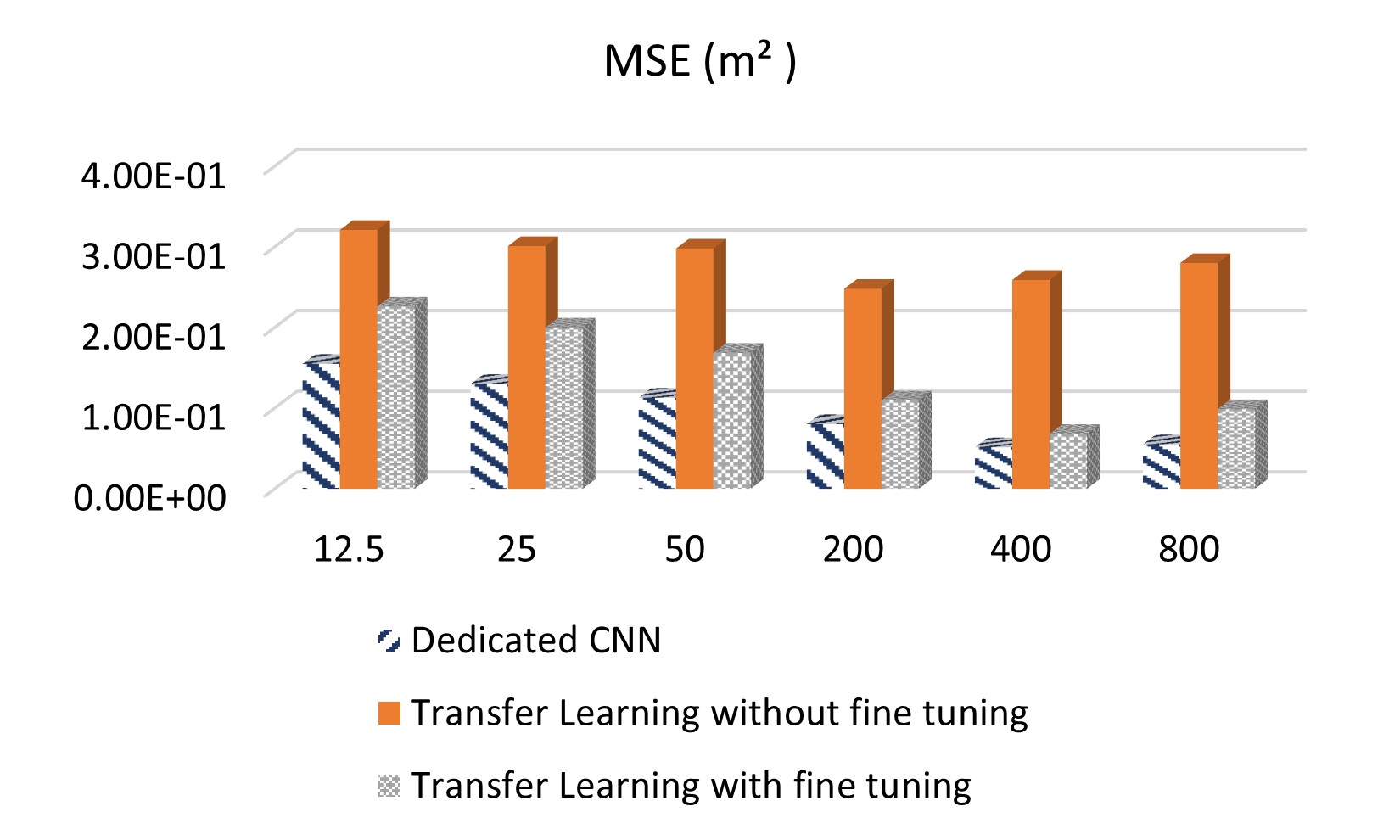}
\caption{MSE comparison between two states of using the transfer learning method and without its usage.}
\label{fig:loss_comp}
\end{center}
\vspace{-7mm}
\end{figure}

\begin{figure}[htbp]

\begin{center}
\includegraphics[scale=0.7]{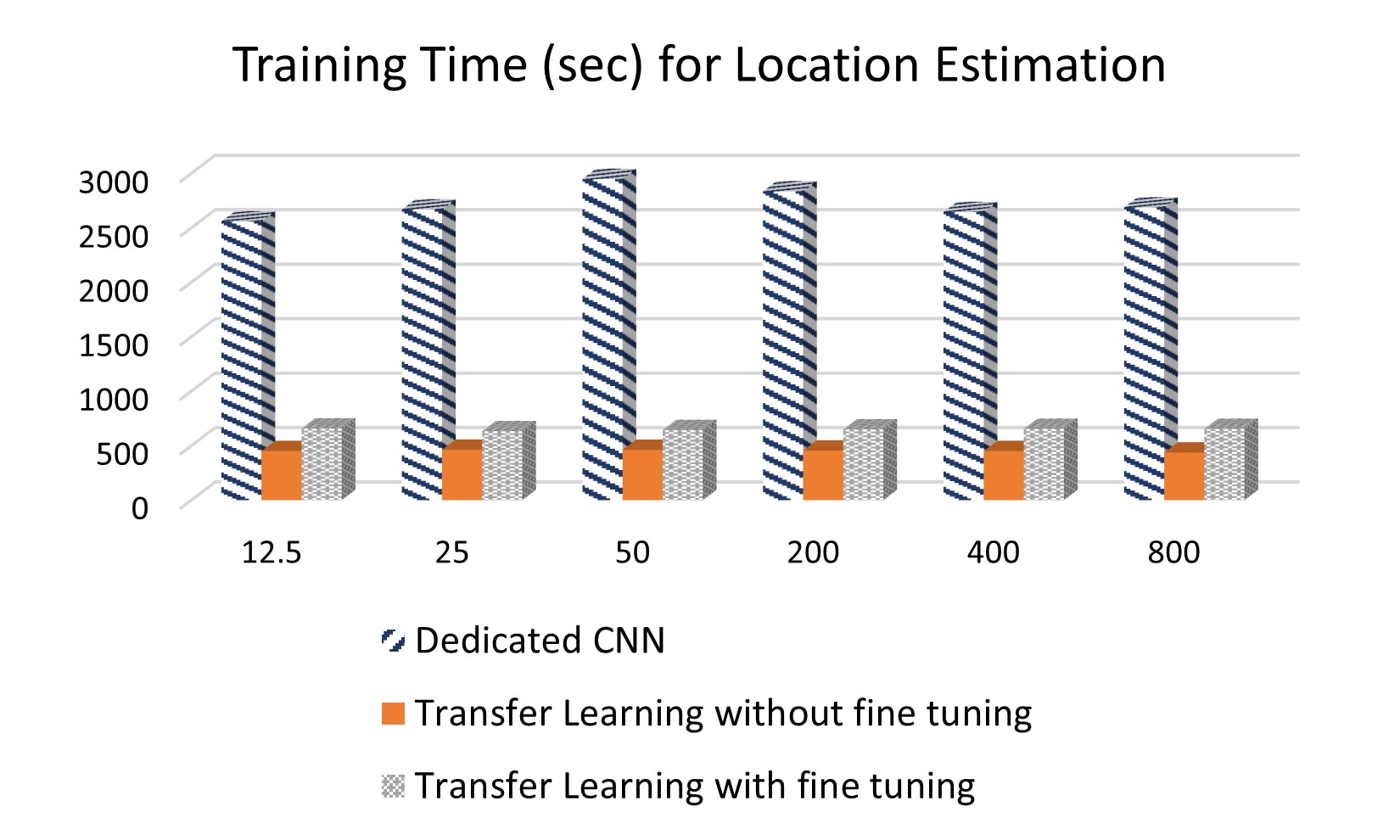}
\caption{Training time comparison between two states of using the transfer learning method and without its usage for location estimation of TL faults.}
\label{fig:train_location}
\end{center}
\vspace{-5mm}
\end{figure}

\begin{figure*}[t]
\centering
\subfigure[]{\includegraphics[width=0.32\textwidth]{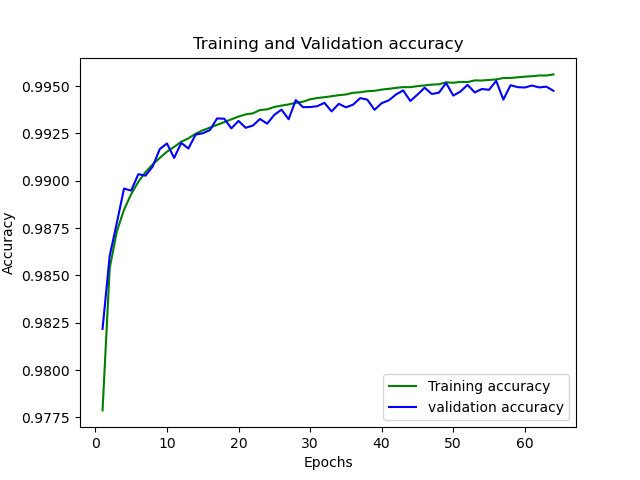}}
\subfigure[]{\includegraphics[width=0.32\textwidth]{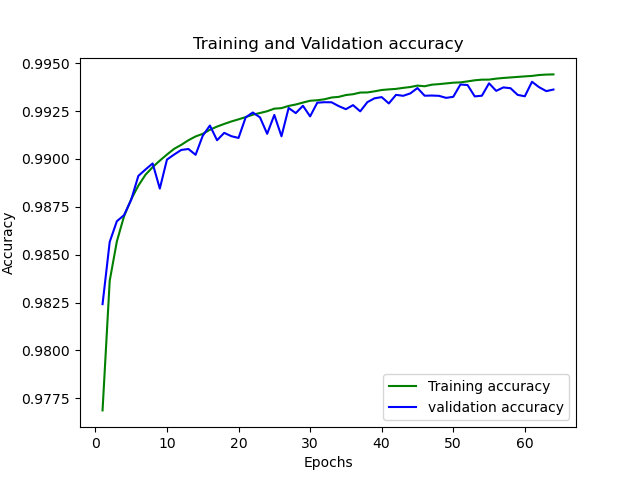}}
\subfigure[]{\includegraphics[width=0.32\textwidth]{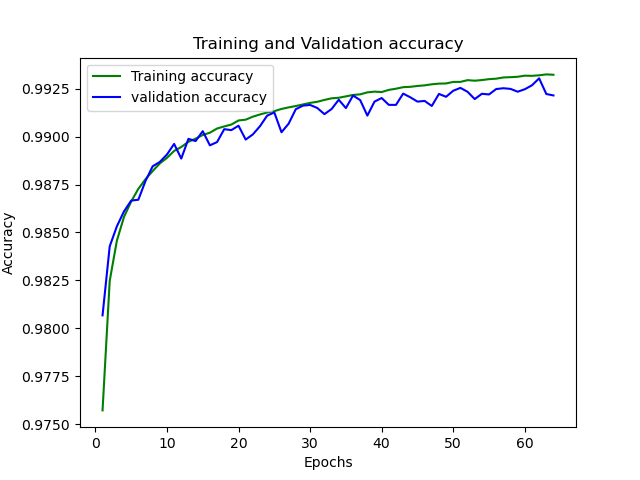}}
\subfigure[]{\includegraphics[width=0.32\textwidth]{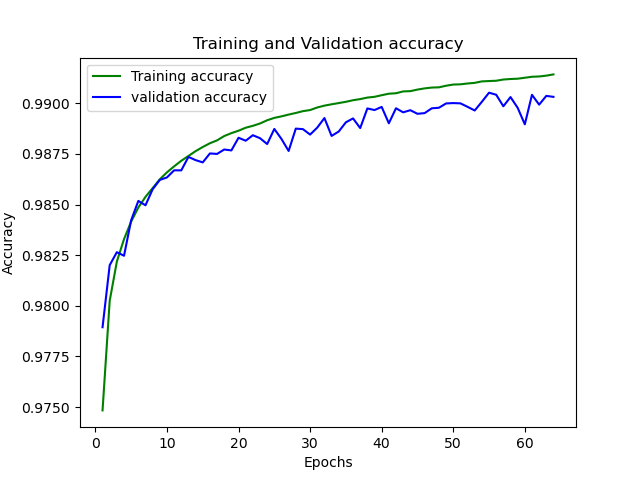}}
\subfigure[]{\includegraphics[width=0.32\textwidth]{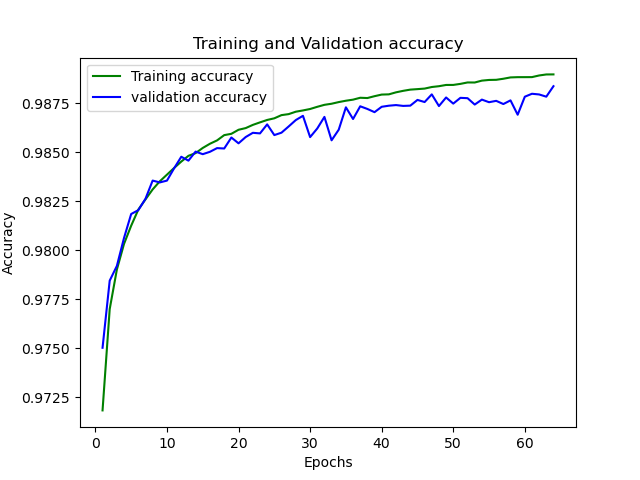}} 
\subfigure[]{\includegraphics[width=0.32\textwidth]{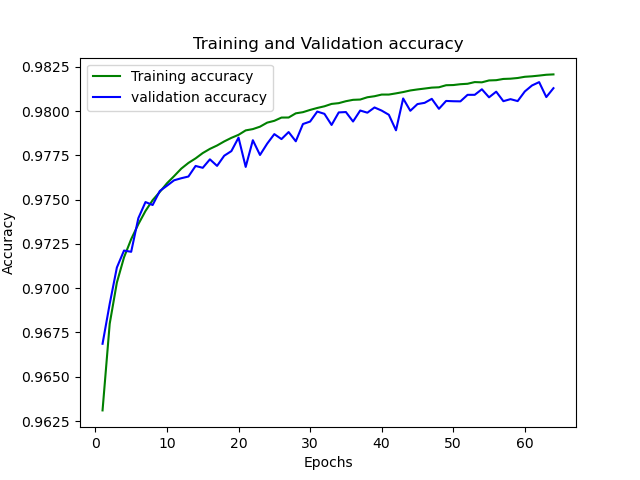}}
\caption{Training and validation accuracy of TLs with the lengths of}{ a. L/8=12.5 $(km)$ b. L/4=25 $(km)$ c. L/2=50 $(km)$ \\d. 2L=200 $(km)$
e. 4L=400 $(km)$ f. 8L=800 $(km)$ based on LeNet-5 using transfer learning method with fine tuning for classification of TL faults.}
\label{fig:acc_6}
\vspace{-1mm}
\end{figure*}

%\vspace{1.2cm}

\subsection{Fault Location Estimation}
To evaluate the accuracy of location estimation methodologies, the mean square error (MSE) which is defines as (\ref{eq:mse}), is calculated for each length of TL individually in the three considered approaches: A dedicated CNN for each length variant of TLs, and with and without the fine tuning process in transfer learning-based technique. The number of epochs is 32 for all of these approaches. 
\begin{equation}
    \label{eq:mse}
    MSE=\sum_{i=1}^{n}(x_i-y_i)^2
\end{equation}

Tables \ref{tab:without_loss}, \ref{tab:loss_nofine}, and \ref{tab:with_loss} indicate the MSE values and the training times of these three approaches in locating the TL faults. Based on these tables, it can be concluded that the proposed transfer learning method decreases the training time of faults location estimation to \underline{one-fourth} of the training time of the dedicated CNN approach (without transfer learning). Furthermore, although the fine tuning process takes negligible amount of time, it reduces the overall error considerably. Based on Fig. \ref{fig:loss_comp}, the MSE differences between these two methodologies (dedicated CNN and transfer learning technique with fine tuning) for each length variant is less than 0.1 ($m^2$) which proves the reliability of the proposed method. One general conclusion that can be made from Figs. \ref{fig:acc_comp} and \ref{fig:loss_comp} is that as the difference between the TL lengths of the source and target datasets increases, the transfer learning-based accuracy  decreases and the validation loss for location estimation increases which is expected considering the feature differences. Fig. \ref{fig:train_location} shows that transfer learning with fine tuning method decreases the training time of the TL fault location estimation to less than \underline{one-fourth} of the state where a dedicated CNN is applied.

\begin{table}[h]
\center
%\vspace{-2mm}
\caption{Location estimation results for TL length variants using dedicated LeNet-5 NN.}
\label{tab:without_loss}
\begin{tabular}{|c|c|c|c|}
\hline
\begin{tabular}[c]{@{}c@{}}Length $(km)$ \end{tabular} & \begin{tabular}[c]{@{}c@{}}\# Epochs\end{tabular} & \begin{tabular}[c]{@{}c@{}}MSE ($m^2$)\end{tabular} & \begin{tabular}[c]{@{}c@{}}Training Time ($sec$)\end{tabular} \\ \hline
12.5&& 1.55$\times$$10^-1$& 2687.36\\ \cline{1-1} \cline{3-4}
25&& 1.03$\times$$10^-1$& 2645.50\\ \cline{1-1} \cline{3-4} 
50&& 1.13$\times$$10^-1$& 2948.40\\ \cline{1-1} \cline{3-4}
100& \multirow{1}{*}{32}& 8.73 $\times$$10^-2$ & 2842.66\\ \cline{1-1} \cline{3-4}
200&& 8.03$\times$$10^-2$& 2835.40\\ \cline{1-1} \cline{3-4}
400&& 5.15$\times$$10^-2$& 2668.42\\ \cline{1-1} \cline{3-4} 
800&& 5.52$\times$$10^-2$& 2558.84\\ \hline
\end{tabular}
\end{table}

\begin{table}[h]
\center

\caption{Location estimation results for TL length variants using LeNet-5-based transfer learning \underline{without} fine tuning.}
\label{tab:loss_nofine}
\begin{tabular}{|c|c|c|c|}
\hline
\begin{tabular}[c]{@{}c@{}}Length $(km)$\end{tabular} & \begin{tabular}[c]{@{}c@{}}\# Epochs\end{tabular} & MSE ($m^2$) & Training Time ($sec$) \\ \hline
12.5& \multirow{6}{*}{32}& 3.21$\times 10^-1$& 550.86\\ \cline{1-1} \cline{3-4} 
25&& 3.01$\times 10^-1$& 562.62\\ \cline{1-1} \cline{3-4} 
50&& 2.98$\times 10^-1$& 561.61\\ \cline{1-1} \cline{3-4} 
200&& 2.48$\times 10^-1$& 555.35\\ \cline{1-1} \cline{3-4} 
400&& 2.59$\times 10^-1$& 550.65\\ \cline{1-1} \cline{3-4} 
800&& 2.80$\times 10^-1$& 537.17\\ \hline
\end{tabular}
\end{table}

\begin{table}[h]
\centering
\caption{Location estimation results for TL length variants using LeNet-5-based transfer learning method \underline{with} fine tuning.}
\label{tab:with_loss}
\begin{tabular}{|c|c|c|c|}
\hline
\begin{tabular}[c]{@{}c@{}}Length $(km)$\end{tabular} & \begin{tabular}[c]{@{}c@{}}\# Epochs\end{tabular} & \begin{tabular}[c]{@{}c@{}}MSE ($m^2$)\end{tabular} & \begin{tabular}[c]{@{}c@{}}Training Time ($sec$)\end{tabular} \\ \hline
12.5&\multirow{6}{*}{32}&2.26$\times 10^-1$& 654.99\\ \cline{1-1} \cline{3-4} 
25&& 2.00$\times 10^-1$& 654.45\\ \cline{1-1} \cline{3-4} 
50& & 1.69$\times 10^-1$& 643.90\\ \cline{1-1} \cline{3-4} 
200&& 1.10$\times 10^-1$& 649.84\\ \cline{1-1} \cline{3-4} 
400&&6.79$\times 10^-2$& 632.82\\ \cline{1-1} \cline{3-4} 
800&& 9.88$\times 10^-2$& 658.35\\ \hline
\end{tabular}
\vspace{-2mm}

\end{table}

Considering the test time of detection, classification, and location estimation of TL faults, it is concluded that the cumulative test time does not exceed 6 $(\mu sec)$ which satisfies the IEEE standard specifications \cite{raza2020review}.

\begin{figure*}[t]
\centering
\subfigure[]{\includegraphics[width=0.32\textwidth]{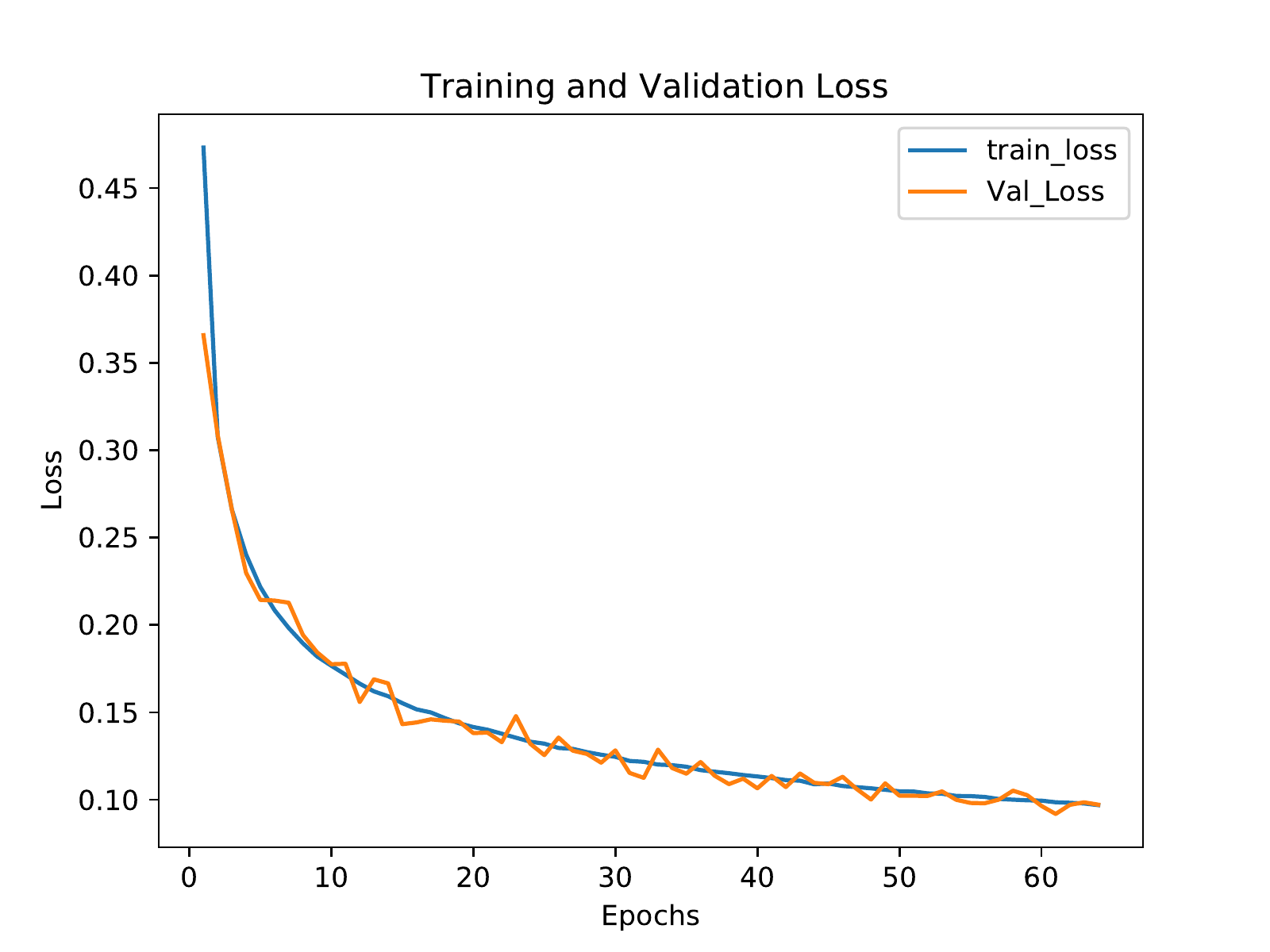}}
\subfigure[]{\includegraphics[width=0.32\textwidth]{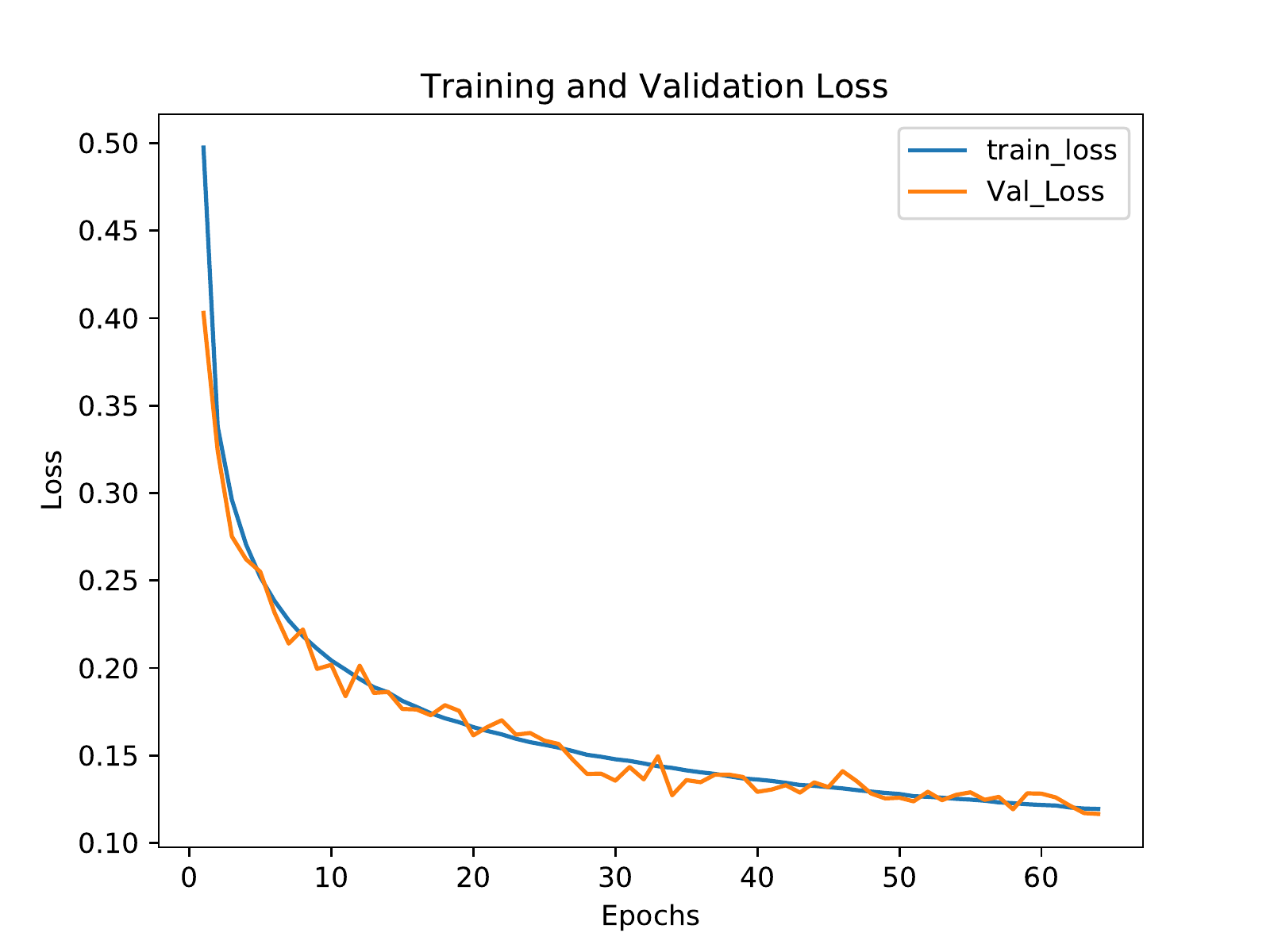}}
\subfigure[]{\includegraphics[width=0.32\textwidth]{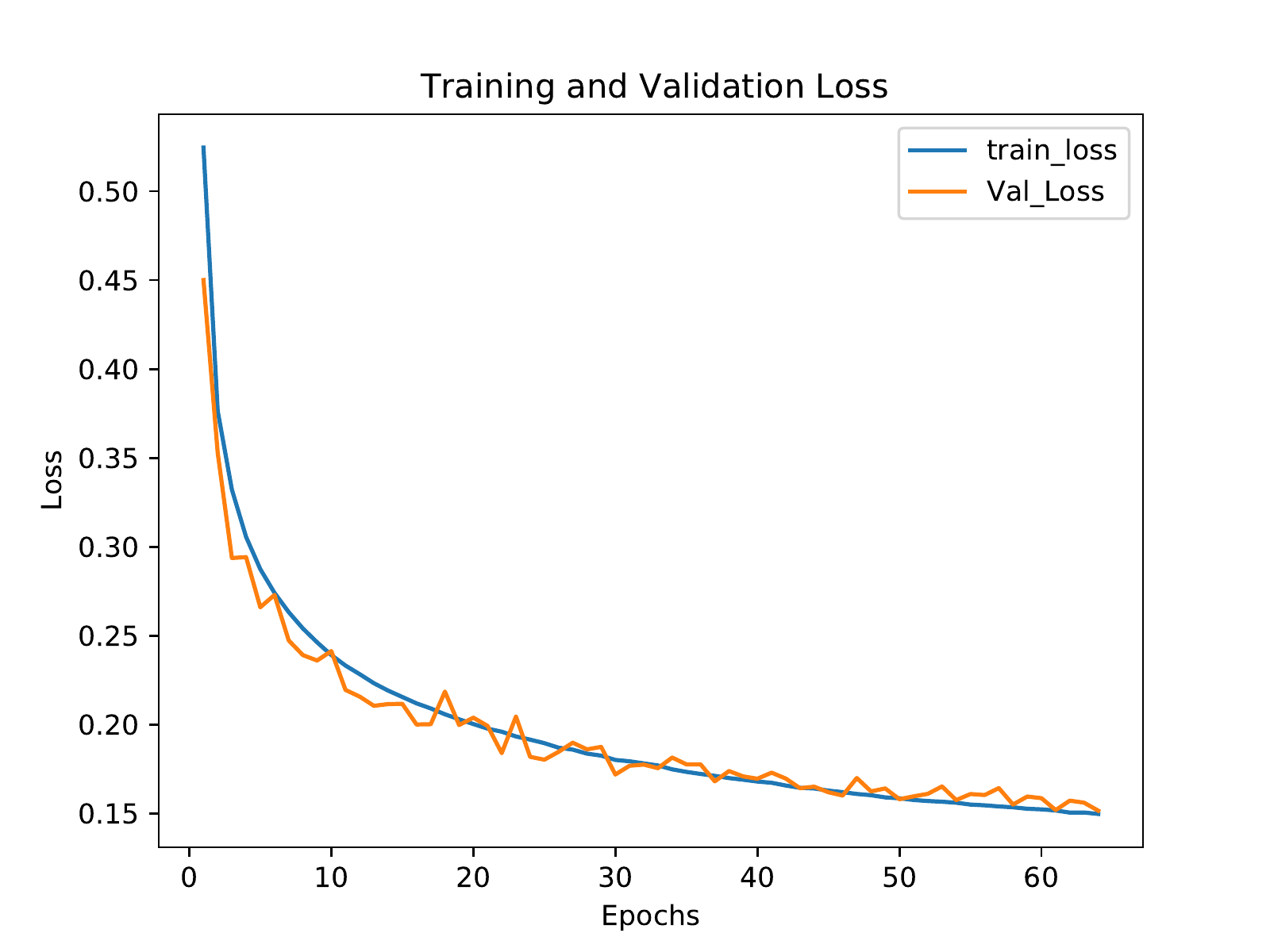}}
\subfigure[]{\includegraphics[width=0.32\textwidth]{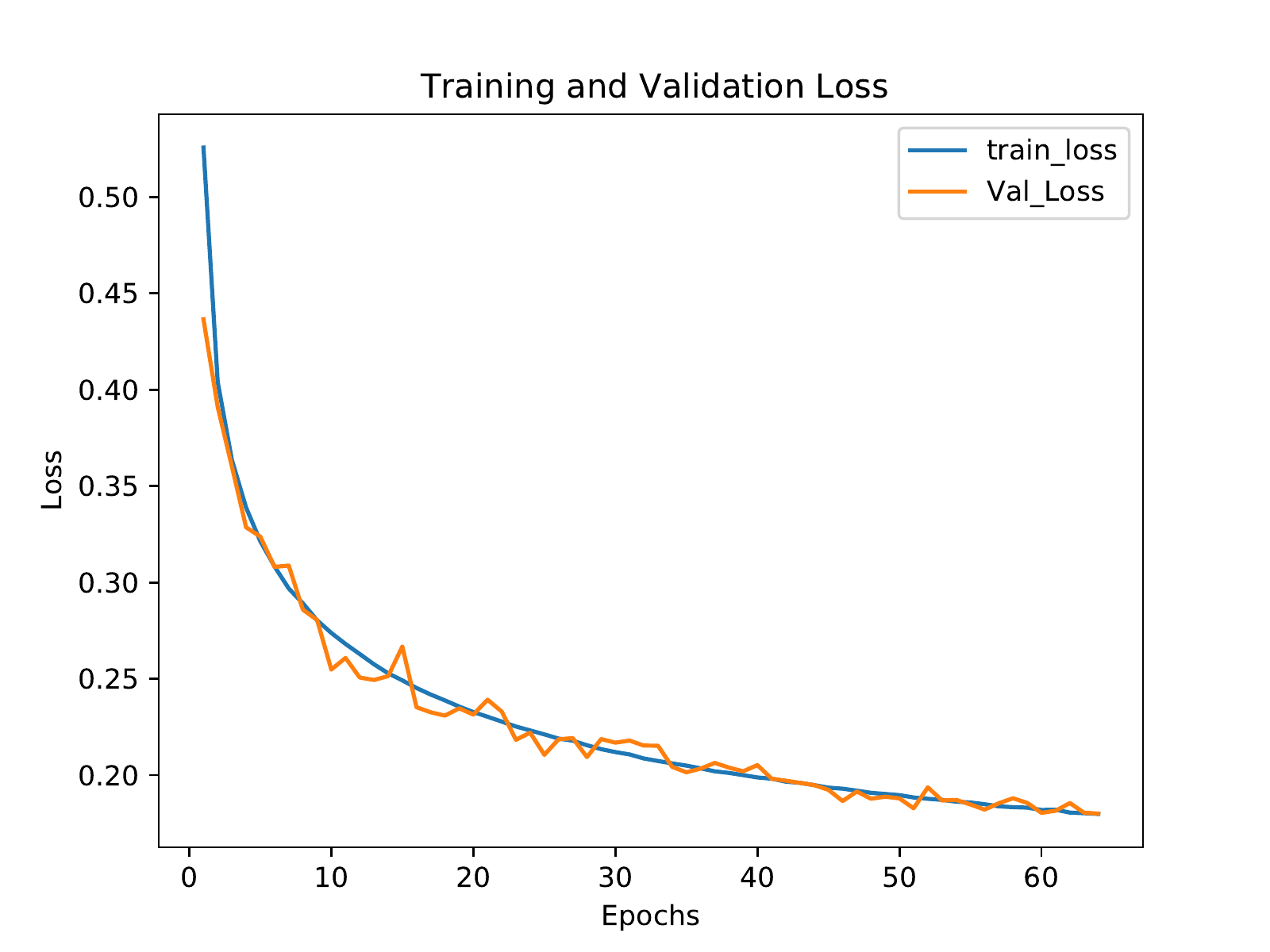}}
\subfigure[]{\includegraphics[width=0.32\textwidth]{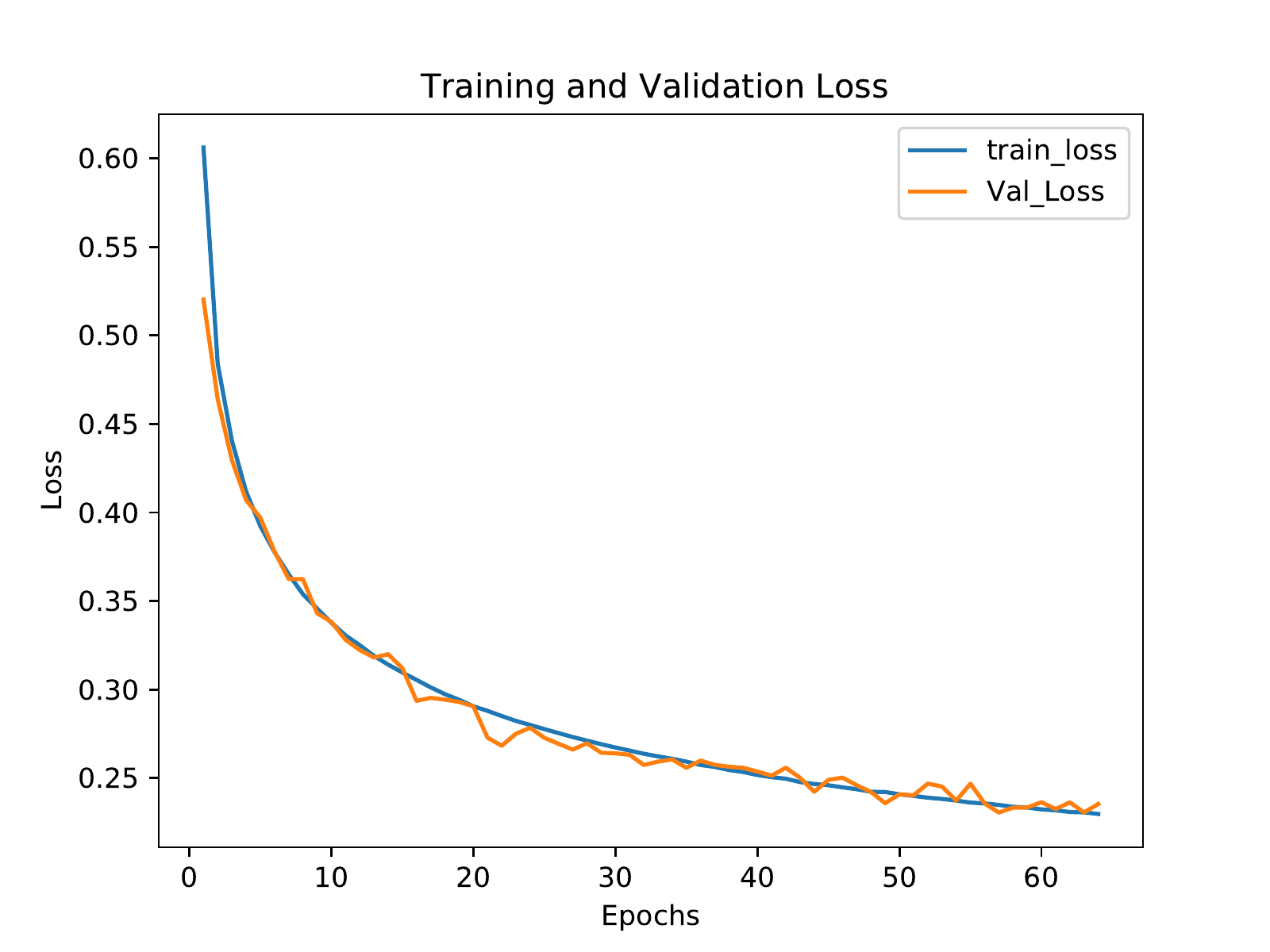}} 
\subfigure[]{\includegraphics[width=0.32\textwidth]{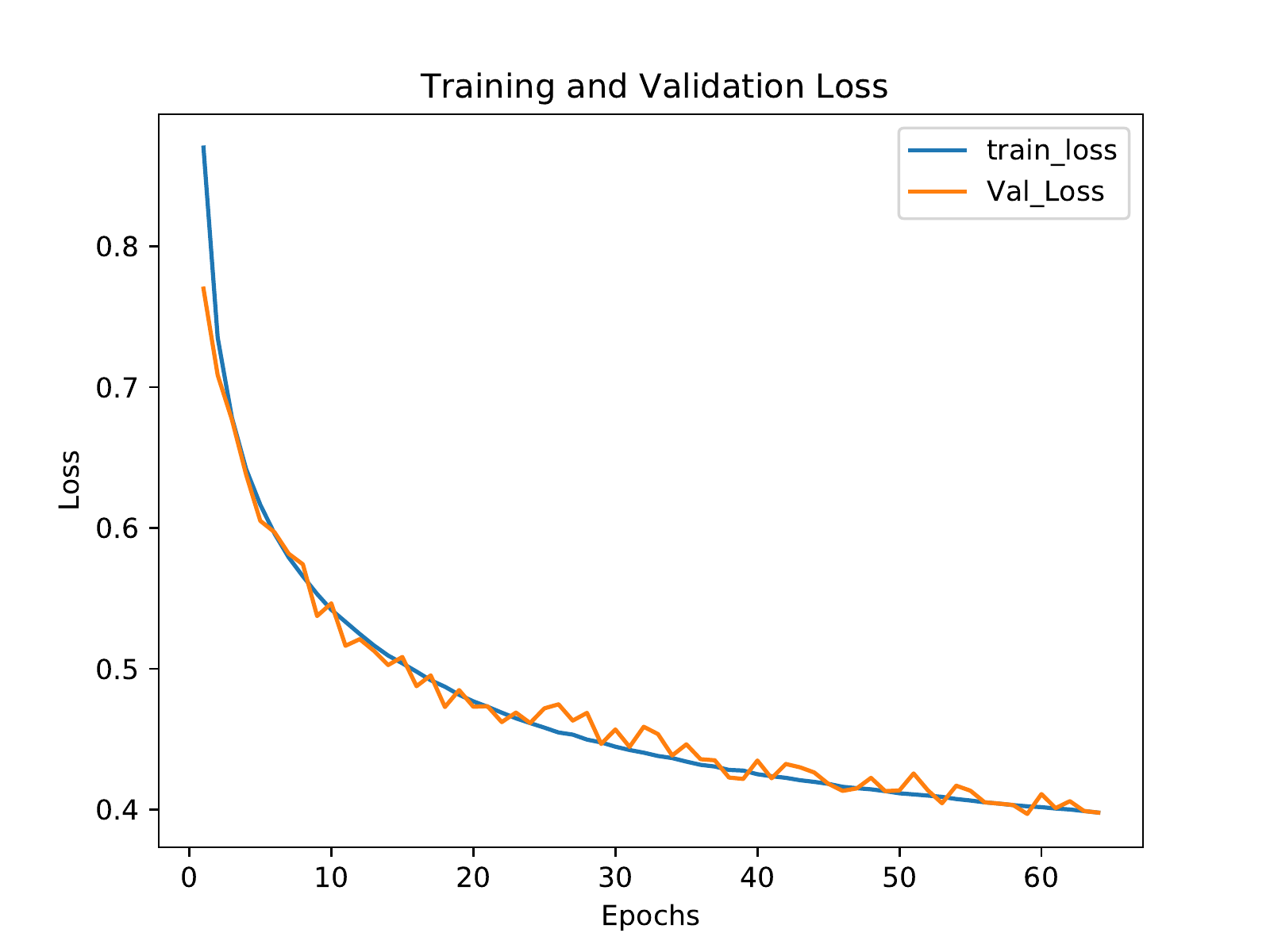}}
\caption{Training and Validation loss of TLs with the lengths of}{ a. L/8=12.5 $(km)$ b. L/4=25 $(km)$ c. L/2=50 $(km)$ \\d. 2L=200 $(km)$ e. 4L=400 $(km)$ f. 8L=800 $(km)$ based on LeNet-5 using transfer learning method with fine tuning for location estimation of TL faults.}
\label{fig:loss_6}
\end{figure*}

Figs. \ref{fig:acc_6} shows the variations of accuracy for 6 different TLs using the transfer learning methodology, and compares the training and validation accuracy values for the detection and classification steps. In Fig. \ref{fig:loss_6}, the loss variations of TLs for training and validation steps are shown for all 6 variants of the TLs. These figures prove the reliability and efficiency of the proposed approach in location estimation of TL faults.

\vspace{-8mm}
\section{Conclusion}\label{conclusion}
%The need for a general solution that is low cost, fast, reliable, and compatible with more than one type of TLs has become viral more than before.
In this paper, a generalized, high-speed, and accurate approach based on the transfer learning methodology is proposed for the first time to diagnose TL faults and estimate their locations. This approach makes use of a pretrained CNN called LeNet-5 which is trained based on the dataset of a TL with the length of $L=100 (km)$, and then the weights of feature extractor layers are frozen and transferred to a new similar CNN. In the second CNN, the fully connected layers update their weights based on the new datasets specific to each length variant of TLs to produce more accurate results. The simulation results show that the training time of the transfer learning-based approach is \underline{half} of the required training time of the dedicated CNN (without using transfer learning) approach for the classification of TL faults, and for the location estimation of TL faults the training time of the proposed method is  \underline{one-fourth} of the required training time of the dedicated CNN (without using transfer learning).
Such remarkable reduction in training time proves the reduction in computational load and therefore memory usage, which have been always significant issues for the associated online datasets. Moreover, the accuracy of the fault classification and the MSE of the fault location estimation are nearly similar to those given by a specifically trained (dedicated) CNN for each length of TLs. These results prove the generality and reliability of the proposed transfer learning methodology for the TL fault diagnosis problems.


\bibliographystyle{model1-num-names}
%\bibliographystyle{elsarticle-num}
%\bibliographystyle{cas-model2-names}
%\bibliographystyle{ieeetr}

% \bibliography{cas-refs}

\begin{thebibliography}{33}
\expandafter\ifx\csname natexlab\endcsname\relax\def\natexlab#1{#1}\fi
\providecommand{\url}[1]{\texttt{#1}}
\providecommand{\href}[2]{#2}
\providecommand{\path}[1]{#1}
\providecommand{\DOIprefix}{doi:}
\providecommand{\ArXivprefix}{arXiv:}
\providecommand{\URLprefix}{URL: }
\providecommand{\Pubmedprefix}{pmid:}
\providecommand{\doi}[1]{\href{http://dx.doi.org/#1}{\path{#1}}}
\providecommand{\Pubmed}[1]{\href{pmid:#1}{\path{#1}}}
\providecommand{\bibinfo}[2]{#2}
\ifx\xfnm\relax \def\xfnm[#1]{\unskip,\space#1}\fi
%Type = Article
\bibitem[{Hare~et al.(2016)}]{hare2016fault}
\bibinfo{author}{J.~Hare~et al.},
\newblock \bibinfo{title}{Fault diagnostics in smart micro-grids: A survey},
\newblock \bibinfo{journal}{Renewable \& Sustainable Energy Reviews}
  \bibinfo{volume}{60} (\bibinfo{year}{2016}) \bibinfo{pages}{1114--1124}.
%Type = Article
\bibitem[{Bjelić et~al.(2022)Bjelić, Brković, Žarković, and
  Miljković}]{BJELIC2022107825}
\bibinfo{author}{M.~Bjelić}, \bibinfo{author}{B.~Brković},
  \bibinfo{author}{M.~Žarković}, \bibinfo{author}{T.~Miljković},
\newblock \bibinfo{title}{Fault detection in a power transformer based on
  reverberation time},
\newblock \bibinfo{journal}{International Journal of Electrical Power \& Energy
  Systems} \bibinfo{volume}{137} (\bibinfo{year}{2022})
  \bibinfo{pages}{107825}.
%Type = Article
\bibitem[{Khoshbouy et~al.(2022)Khoshbouy, Yazdaninejadi, and
  Bolandi}]{KHOSHBOUY2022107826}
\bibinfo{author}{M.~Khoshbouy}, \bibinfo{author}{A.~Yazdaninejadi},
  \bibinfo{author}{T.~G. Bolandi},
\newblock \bibinfo{title}{Transmission line adaptive protection scheme: A new
  fault detection approach based on pilot superimposed impedance},
\newblock \bibinfo{journal}{International Journal of Electrical Power \& Energy
  Systems} \bibinfo{volume}{137} (\bibinfo{year}{2022})
  \bibinfo{pages}{107826}.
%Type = Article
\bibitem[{Zheng et~al.(2022)Zheng, Li, Xu, Kong, Wang, Lin, and
  Zhang}]{ZHENG2022107658}
\bibinfo{author}{J.~Zheng}, \bibinfo{author}{P.~Li}, \bibinfo{author}{K.~Xu},
  \bibinfo{author}{X.~Kong}, \bibinfo{author}{C.~Wang},
  \bibinfo{author}{J.~Lin}, \bibinfo{author}{C.~Zhang},
\newblock \bibinfo{title}{A distance protection scheme for hvdc transmission
  lines based on the steady-state parameter model},
\newblock \bibinfo{journal}{International Journal of Electrical Power \& Energy
  Systems} \bibinfo{volume}{136} (\bibinfo{year}{2022})
  \bibinfo{pages}{107658}.
%Type = Article
\bibitem[{Fahim~et al.(2020)}]{fahim2020self}
\bibinfo{author}{S.~R. Fahim~et al.},
\newblock \bibinfo{title}{Self attention convolutional neural network with time
  series imaging based feature extraction for transmission line fault detection
  \& classification},
\newblock \bibinfo{journal}{Electric Power Sys. Research} \bibinfo{volume}{187}
  (\bibinfo{year}{2020}) \bibinfo{pages}{106437}.
%Type = Article
\bibitem[{Rai~et al.(2020)}]{rai2020fault}
\bibinfo{author}{P.~Rai~et al.},
\newblock \bibinfo{title}{Fault classification in power system distribution
  network integrated with distributed generators using cnn},
\newblock \bibinfo{journal}{Electric Power Sys. Research}
  (\bibinfo{year}{2020}) \bibinfo{pages}{106914}.
%Type = Article
\bibitem[{Fuada~et al.(2020)}]{fuada2020high}
\bibinfo{author}{S.~Fuada~et al.},
\newblock \bibinfo{title}{A high-accuracy of transmission line faults (tlfs)
  classification based on convolutional neural network},
\newblock \bibinfo{journal}{Intl. J. of Electronics \& Telecommunications}
  \bibinfo{volume}{66} (\bibinfo{year}{2020}) \bibinfo{pages}{655--664}.
%Type = Article
\bibitem[{Shiddieqy~et al.(2019)}]{shiddieqy2019power}
\bibinfo{author}{H.~Shiddieqy~et al.},
\newblock \bibinfo{title}{Power line transmission fault modeling \& dataset
  generation for ai based automatic detection},
\newblock \bibinfo{journal}{Intl. Symposium on Electronics \& Smart Devices
  (ISESD)}  (\bibinfo{year}{2019}) \bibinfo{pages}{1--5}.
%Type = Article
\bibitem[{Chen~et al.(2016)}]{chen2016detection}
\bibinfo{author}{K.~Chen~et al.},
\newblock \bibinfo{title}{Detection \& classification of transmission line
  faults based on unsupervised feature learning \& convolutional sparse
  autoencoder},
\newblock \bibinfo{journal}{IEEE Tran. on Smart Grid} \bibinfo{volume}{9}
  (\bibinfo{year}{2016}) \bibinfo{pages}{1748--1758}.
%Type = Article
\bibitem[{Fawaz et~al.(2018)Fawaz, Forestier, Weber, Idoumghar, and
  Muller}]{fawaz2018transfer}
\bibinfo{author}{H.~I. Fawaz}, \bibinfo{author}{G.~Forestier},
  \bibinfo{author}{J.~Weber}, \bibinfo{author}{L.~Idoumghar},
  \bibinfo{author}{P.-A. Muller},
\newblock \bibinfo{title}{Transfer learning for time series classification},
\newblock \bibinfo{journal}{IEEE international conference on big data (Big
  Data)}  (\bibinfo{year}{2018}) \bibinfo{pages}{1367--1376}.
%Type = Article
\bibitem[{Guo~et al.(2018)}]{guo2018deep}
\bibinfo{author}{L.~Guo~et al.},
\newblock \bibinfo{title}{Deep convolutional transfer learning network: A new
  method for intelligent fault diagnosis of machines with unlabeled data},
\newblock \bibinfo{journal}{IEEE Trans. on Industrial Electronics}
  \bibinfo{volume}{66} (\bibinfo{year}{2018}) \bibinfo{pages}{7316--7325}.
%Type = Article
\bibitem[{Shao~et al.(2018)}]{shao2018highly}
\bibinfo{author}{S.~Shao~et al.},
\newblock \bibinfo{title}{Highly accurate machine fault diagnosis using deep
  transfer learning},
\newblock \bibinfo{journal}{IEEE Transactions on Industrial Informatics}
  \bibinfo{volume}{15} (\bibinfo{year}{2018}) \bibinfo{pages}{2446--2455}.
%Type = Article
\bibitem[{Shao~et al.(2020)}]{shao2020intelligent}
\bibinfo{author}{H.~Shao~et al.},
\newblock \bibinfo{title}{Intelligent fault diagnosis of rotor-bearing system
  under varying working conditions with modified transfer convolutional neural
  network and thermal images},
\newblock \bibinfo{journal}{IEEE Transactions on Industrial Informatics}
  \bibinfo{volume}{17} (\bibinfo{year}{2020}) \bibinfo{pages}{3488--3496}.
%Type = Article
\bibitem[{Xu~et al.(2019)}]{xu2019digital}
\bibinfo{author}{Y.~Xu~et al.},
\newblock \bibinfo{title}{A digital-twin-assisted fault diagnosis using deep
  transfer learning},
\newblock \bibinfo{journal}{IEEE Access} \bibinfo{volume}{7}
  (\bibinfo{year}{2019}) \bibinfo{pages}{19990--19999}.
%Type = Article
\bibitem[{Li et~al.(2020)Li, Huang, He, Wang, Li, and Li}]{li2020deep}
\bibinfo{author}{J.~Li}, \bibinfo{author}{R.~Huang}, \bibinfo{author}{G.~He},
  \bibinfo{author}{S.~Wang}, \bibinfo{author}{G.~Li}, \bibinfo{author}{W.~Li},
\newblock \bibinfo{title}{A deep adversarial transfer learning network for
  machinery emerging fault detection},
\newblock \bibinfo{journal}{IEEE Sensors J.} \bibinfo{volume}{20}
  (\bibinfo{year}{2020}) \bibinfo{pages}{8413--8422}.
%Type = Article
\bibitem[{Hubel(1962)}]{hubel1962receptive}
\bibinfo{author}{T.~N. Hubel, David H \&~Wiesel},
\newblock \bibinfo{title}{Receptive fields, binocular interaction \& functional
  architecture in the cat's visual cortex},
\newblock \bibinfo{journal}{The J. of physiology} \bibinfo{volume}{160}
  (\bibinfo{year}{1962}) \bibinfo{pages}{106--154}.
%Type = Article
\bibitem[{Kiruthika(2020)}]{kiruthika2020classification}
\bibinfo{author}{S.~Kiruthika, M \&~Bindu},
\newblock \bibinfo{title}{Classification of electrical power system conditions
  with convolutional neural networks},
\newblock \bibinfo{journal}{Engineering, Tech \& Applied Science Research}
  \bibinfo{volume}{10} (\bibinfo{year}{2020}) \bibinfo{pages}{5759--5768}.
%Type = Article
\bibitem[{Lei(2019)}]{lei2019intelligent}
\bibinfo{author}{Z.~Lei, Xusheng \&~Sui},
\newblock \bibinfo{title}{Intelligent fault detection of high voltage line
  based on the faster r-cnn},
\newblock \bibinfo{journal}{Measurement} \bibinfo{volume}{138}
  (\bibinfo{year}{2019}) \bibinfo{pages}{379--385}.
%Type = Article
\bibitem[{Wang~et al.(2019)}]{wang2019image}
\bibinfo{author}{Y.~Wang~et al.},
\newblock \bibinfo{title}{Image classification towards transmission line fault
  detection via learning deep quality-aware fine-grained categorization},
\newblock \bibinfo{journal}{J. of Visual Comm \& Image Rep}
  \bibinfo{volume}{64} (\bibinfo{year}{2019}) \bibinfo{pages}{102647}.
%Type = Article
\bibitem[{Dai~et al.(2020)}]{dai2020fast}
\bibinfo{author}{Z.~Dai~et al.},
\newblock \bibinfo{title}{Fast \& accurate cable detection using cnn},
\newblock \bibinfo{journal}{Applied Intelligence} \bibinfo{volume}{50}
  (\bibinfo{year}{2020}) \bibinfo{pages}{4688--4707}.
%Type = Article
\bibitem[{Pan and Yang(2009)}]{pan2009survey}
\bibinfo{author}{S.~J. Pan}, \bibinfo{author}{Q.~Yang},
\newblock \bibinfo{title}{A survey on transfer learning},
\newblock \bibinfo{journal}{IEEE Transactions on knowledge and data
  engineering} \bibinfo{volume}{22} (\bibinfo{year}{2009})
  \bibinfo{pages}{1345--1359}.
%Type = Article
\bibitem[{Zheng~et al.(2019)}]{zheng2019cross}
\bibinfo{author}{H.~Zheng~et al.},
\newblock \bibinfo{title}{Cross-domain fault diagnosis using knowledge transfer
  strategy: a review},
\newblock \bibinfo{journal}{IEEE Access} \bibinfo{volume}{7}
  (\bibinfo{year}{2019}) \bibinfo{pages}{129260--129290}.
%Type = Article
\bibitem[{Csurka(2017)}]{csurka2017domain}
\bibinfo{author}{G.~Csurka},
\newblock \bibinfo{title}{Domain adaptation for visual applications: A
  comprehensive survey},
\newblock \bibinfo{journal}{arXiv preprint arXiv:1702.05374}
  (\bibinfo{year}{2017}).
%Type = Article
\bibitem[{UDOFIA~et al.(2020)}]{udofia2020fault}
\bibinfo{author}{K.~M. UDOFIA~et al.},
\newblock \bibinfo{title}{Fault detection, classification \& location on 132kv
  transmission line based on dwt \& anfis},
\newblock \bibinfo{journal}{surge (VS)} \bibinfo{volume}{7}
  (\bibinfo{year}{2020}).
%Type = Article
\bibitem[{AbdelAziz(2016)}]{abdel2016detection}
\bibinfo{author}{.~D. AbdelAziz, \&~Hasaneen},
\newblock \bibinfo{title}{Detection \& classification of one conductor open
  faults in parallel transmission line using artificial neural network},
\newblock \bibinfo{journal}{Intl. J. of Scientific Research \& Engineering
  Trends} \bibinfo{volume}{2} (\bibinfo{year}{2016}) \bibinfo{pages}{139--146}.
%Type = Article
\bibitem[{Mahmud~et al.(2018)}]{mahmud2018robust}
\bibinfo{author}{M.~N. Mahmud~et al.},
\newblock \bibinfo{title}{A robust transmission line fault classification
  scheme using class-dependent feature \& 2-tier multilayer perceptron
  network},
\newblock \bibinfo{journal}{Electrical Engineering} \bibinfo{volume}{100}
  (\bibinfo{year}{2018}) \bibinfo{pages}{607--623}.
%Type = Article
\bibitem[{LeCun et~al.(2015)}]{lecun2015lenet}
\bibinfo{author}{Y.~LeCun}, et~al.,
\newblock \bibinfo{title}{Lenet-5, convolutional neural networks},
\newblock \bibinfo{journal}{URL: http://yann. lecun. com/exdb/lenet}
  \bibinfo{volume}{20} (\bibinfo{year}{2015}) \bibinfo{pages}{14}.
%Type = Misc
\bibitem[{Chollet et~al.(2015)}]{chollet2015keras}
\bibinfo{author}{F.~Chollet}, et~al., \bibinfo{title}{Keras},
  \bibinfo{year}{2015}. \URLprefix \url{https://github.com/fchollet/keras}.
%Type = Misc
\bibitem[{Abedi~et al.(2015)}]{tensorflow2015-whitepaper}
\bibinfo{author}{M.~Abedi~et al.}, \bibinfo{title}{{TensorFlow}: Large-scale
  machine learning on heterogeneous systems}, \bibinfo{year}{2015}. \URLprefix
  \url{https://www.tensorflow.org/}, \bibinfo{note}{software available from
  tensorflow.org}.
%Type = Article
\bibitem[{Pedregosa~et al.(2011)}]{pedregosa2011scikit}
\bibinfo{author}{F.~Pedregosa~et al.},
\newblock \bibinfo{title}{Scikit-learn: Machine learning in python},
\newblock \bibinfo{journal}{the Journal of machine Learning research}
  \bibinfo{volume}{12} (\bibinfo{year}{2011}) \bibinfo{pages}{2825--2830}.
%Type = Article
\bibitem[{Haroush~et al.(2021)}]{haroush2021statistical}
\bibinfo{author}{M.~Haroush~et al.},
\newblock \bibinfo{title}{Statistical testing for efficient out of distribution
  detection in deep neural networks},
\newblock \bibinfo{journal}{arXiv preprint arXiv:2102.12967}
  (\bibinfo{year}{2021}).
%Type = Article
\bibitem[{Zhao et~al.(2021)Zhao, Ge, and Chen}]{zhao2021landslide}
\bibinfo{author}{B.~Zhao}, \bibinfo{author}{Y.~Ge}, \bibinfo{author}{H.~Chen},
\newblock \bibinfo{title}{Landslide susceptibility assessment for a
  transmission line in gansu province, china by using a hybrid approach of
  fractal theory, information value, and random forest models},
\newblock \bibinfo{journal}{Environmental Earth Sciences} \bibinfo{volume}{80}
  (\bibinfo{year}{2021}) \bibinfo{pages}{1--23}.
%Type = Article
\bibitem[{Raza~et al.(2020)}]{raza2020review}
\bibinfo{author}{A.~Raza~et al.},
\newblock \bibinfo{title}{A review of fault diagnosing methods in power
  transmission sys.},
\newblock \bibinfo{journal}{Applied Sciences} \bibinfo{volume}{10}
  (\bibinfo{year}{2020}) \bibinfo{pages}{1312}.

\end{thebibliography}

\vspace{10mm}

\end{document}